\pdfoutput=1  % arXiv: build as PDF
\documentclass[10pt,twocolumn]{article}
\usepackage[letterpaper,margin=0.75in,columnsep=0.25in]{geometry}
\usepackage[T1]{fontenc}
\usepackage{textcomp}
\usepackage{mathptmx}
\usepackage{graphicx}
\usepackage{booktabs}
\usepackage{makecell}
\usepackage{multirow}
\usepackage{listings}
\usepackage{xcolor}

\definecolor{mlirkw}{RGB}{28,60,190}
\definecolor{mlirtype}{RGB}{0,110,110}
\definecolor{mlirstr}{RGB}{178,32,32}
\definecolor{mlircmt}{RGB}{0,128,64}
\definecolor{mlirattr}{RGB}{120,40,160}
\definecolor{lnum}{RGB}{140,140,140}
\definecolor{novacol}{HTML}{1C4E63}
\definecolor{ptcol}{HTML}{A6DAF0}
\definecolor{tritoncol}{HTML}{3AA7D3}
\definecolor{novaline}{HTML}{04C1DE}
\definecolor{eagerline}{HTML}{6607F5}
\definecolor{ptline}{HTML}{F5077E}
\newcommand{\swatch}[1]{\textcolor{#1}{\rule[0.12em]{0.95em}{0.62em}}}

\lstdefinelanguage{MLIR}{
  alsoletter={.\_\%\#@},
  morekeywords={Func.func,func.func,return,ins,outs,to,layout,step,
    nova.matmul,nova.vjp,nova.linear,nova.gelu,nova.sce,nova.constant,
    nova.mul,nova.sce\_backward,nova.gelu\_backward,nova.linear\_backward,
    linalg.matmul,linalg.generic,vector.contract,scf.for,
    arith.constant,arith.andi,arith.shli,arith.xori,arith.addf,arith.select,
    arith.subf,arith.mulf,arith.divf,arith.maximumf,arith.cmpi,arith.trunci,
    arith.index\_castui,linalg.index,math.exp,math.exp2,linalg.yield,scf.yield,
    iter\_args,nova.attention,nova.online\_attention,nova.attention\_backward,
    gpu.shuffle,llvm.call,xor,
    nvgpu.device\_async\_copy,nvgpu.device\_async\_create\_group,
    nvgpu.device\_async\_wait,nvgpu.ldmatrix,nvgpu.mma.sync,
    nova\_vector\_ext.to\_layout},
  morekeywords=[2]{tensor,vector,memref,index,f32,i32,i16,bf16,dense},
  morekeywords=[3]{select,exp,attrs,nova.causal,nova.flash\_mma,nova.causal\_clamped,
    \#nova.lowering\_config,\#nested,\#nested1,\#mapA,\#mapB,\#mapC,
    \#vector.kind,lowering\_config,workgroup,reduction,subgroup,wg\_subgroup,
    mma\_kind,promoted\_operands,indexing\_maps,iterator\_types,kind,
    numTiles,transpose,numGroups,value,pad\_operands,thread},
  morecomment=[l]{//},
  morestring=[b]",
  sensitive=true
}

\usepackage[numbers,sort&compress]{natbib}
\usepackage[hidelinks]{hyperref}
\usepackage{url}
\usepackage{float}
\usepackage{dblfloatfix}
\usepackage{microtype}

\newcommand\blfootnote[1]{%
  \begingroup\renewcommand\thefootnote{}\footnote{#1}%
  \addtocounter{footnote}{-1}\endgroup}

\title{\bfseries Nova JIT Compiler}
\author{}
\date{}

\begin{document}

\twocolumn[
  \begin{center}
    \vspace*{0.6em}
    {\LARGE\bfseries Nova: An End-to-End MLIR Compiler for Deep Learning\par}

    \vspace{2.5em}
    {\normalsize
      Adwaid Suresh\hspace{1.2em}Aparna A\hspace{1.2em}Harshini V M\hspace{1.2em}Jona Delcy C A\\[0.55em]
      Killi Uma Maheswara Rao\hspace{1.2em}Ram Charan Golla\hspace{1.2em}Surendra Vendra\par}
    \vspace{1.4em}
    {\large Blubridge AI\par}
    \vspace{0.25em}
    {\normalsize\ttfamily research@blubridge.com\par}
  \end{center}
  \vspace{2.0em}
]

\noindent\textbf{\textit{Abstract:}} \textit{The performance of deep learning models at scale relies heavily on how effectively high-level mathematical operations are mapped to underlying physical hardware. While high-level tensor frameworks provide flexible abstractions, their execution models inherently lack the whole-graph visibility required to maximize hardware utilization---often forcing a reliance on opaque, hand-written kernel libraries for complex operations like Attention. To bridge this gap, we present the next iteration of Nova, an automated end-to-end JIT compiler that achieves absolute control over hardware mapping by synthesizing fine-grained kernels directly from the computation's structure. In this work, we extend Nova's compilation pipeline to natively support full Transformer architectures. By capturing eager executions and unifying forward and backward passes into a single value-semantic dialect, Nova unlocks aggressive whole-graph optimizations. Rather than relying on rigid, pre-compiled library calls, Nova focuses on extensive cross-operator fusions---collapsing complex causal attention sub-graphs, element-wise operations, and memory-bound normalizations directly into single fused kernels to drastically reduce global memory roundtrips. In our evaluations training a full GPT-2 architecture on Ada 6000 GPUs, Nova demonstrates superior end-to-end throughput, averaging 441K tokens/second compared to 406K for our own eager execution and 405K for \texttt{torch.compile}. By drastically reducing memory-bound overheads through compiler-native fusion, Nova enables efficient full LLM compilation on modern hardware while strictly maintaining numerical parity.}
\blfootnote{All results were produced with LLVM/MLIR 21.1.6 and CUDA Toolkit 13.0.}

\section{Introduction}

The progress of deep learning has been carried in large part by the steady rise in the compute that parallel hardware, and GPUs in particular, can deliver. But that compute is only useful to the extent a model's computation is actually mapped onto it well. This is felt most acutely in training, where the cost of deep learning concentrates: a single training run can occupy a GPU for days, and how much of that time is spent productively depends far less on how the model is written than on how its operations are laid out, kept in memory, and scheduled on the device. The quality of that mapping, not the mathematics of the model, is what sets the performance a model achieves, and producing it still relies heavily on manual effort.

Today, that mapping is delivered chiefly by the specialized kernel libraries that major frameworks rely on (e.g., cuDNN~\cite{cudnn}, cuBLAS, or hand-tuned FlashAttention). These libraries provide carefully optimized implementations of a limited set of standard operations, such as matrix multiplications and self-attention. While these individual pre-compiled operations run close to the device's theoretical limits, this reliance introduces a hard optimization boundary. Because eager frameworks must treat these library calls as opaque black boxes, intermediate data must be written back to global memory between each step. As a result, critical opportunities to aggressively fuse surrounding computation directly into the core matrix operations are fundamentally lost, inflating global memory traffic~\cite{datamovement}. Furthermore, anything outside the library's predefined set reaches only a fraction of the attainable throughput. A novel operation introduced by research falls off the fast path, leaving most of the device idle. Moreover, because such hand-tuned implementations carry rigid assumptions about a particular architecture, these fast paths must be entirely rebuilt for each new hardware generation.

Compilation is the natural way past a fixed library: rather than selecting from pre-built kernels, a compiler generates code for the computation at hand. Existing deep-learning compilers take this route~\cite{iree,openxla,halide,tc,tvm}, yet in practice each remains partly tied to the machinery it was meant to supersede, falling back to hand-optimized kernels on unsupported cases, leaning on expensive autotuning search, or handing off to a separate hardware-specific toolchain.

Nova is an end-to-end JIT compiler built to break this dependence. Operating as the backend for the BluTrain framework~\cite{blutrain}, Nova optimizes the computation as a cohesive whole rather than a sequence of isolated operations, synthesizing optimized machine code directly from the structural semantics of the computation graph.

Specifically, this paper makes the following contributions:
\begin{itemize}
  \item \textbf{An End-to-End Training Pipeline:} Nova provides a full end-to-end pipeline that maps high-level model code directly to optimized machine instructions for dynamic training runs, breaking the opaque boundaries of eager execution.
  \item \textbf{The nova Dialect:} To provide the global visibility required for whole-graph optimizations, Nova introduces a hardware and framework agnostic frontend dialect that natively unifies forward and backward passes into a single execution block.
  \item \textbf{Hardware-Aware Memory \& Compute Orchestration:} Nova replaces search-based autotuning with a deterministic lowering pipeline that maps directly to low-level hardware features. Driven by an Analytic Configurator, Nova derives tile sizes, software pipeline depths, and multi-buffering strategies from explicit hardware constraints, and orchestrates device-level optimizations including automated \texttt{ldmatrix} mapping, shared memory swizzling, vectorization, and in-place memory reuse to sustain high Tensor Core utilization.
  \item \textbf{A Low-Overhead JIT Runtime:} To ensure intensive compilation does not bottleneck the dynamic training loop, a strict caching architecture binds the executable to live memory, guaranteeing compilation costs are paid exactly once.
\end{itemize}

Because Nova derives its execution schedules analytically rather than relying on hardcoded heuristics, it provides seamless portability across disparate hardware. We validate this by scaling Nova directly from consumer accelerators (NVIDIA RTX 3060) to massive enterprise architectures (NVIDIA RTX 6000 Ada). By automatically adapting thread-block tiling and MMA schedules to the shifting SM counts and shared memory limits of the target device, the compiler avoids underutilization on smaller GPUs while saturating the Tensor Cores of enterprise hardware. As a result, Nova sustains peak throughput on full Transformer architectures natively, bridging hardware tiers without requiring a single manual kernel rewrite. It matches the performance of heavily hand-tuned FlashAttention baselines while simultaneously lowering the baseline memory footprint.

This technical report details Nova's systems implementation. We first detail the unified differentiable pipeline (Section~\ref{sec:ir}), which encompasses the nova dialect and automated cross-GPU gradient synchronization. We then demonstrate how this representation is mapped to hardware instructions (Section~\ref{sec:backend}), detailing aggressive whole-graph fusion, causal attention lowering, and pipeline vectorization. Finally, we outline the low-overhead runtime binding (Section~\ref{sec:runtime}) before evaluating Nova's performance (Section~\ref{sec:eval}), first at the kernel level (Section~\ref{sec:microbench}) and then on an end-to-end LLM training step (Section~\ref{sec:e2e}).

\section{Overview}

Nova takes a model written as ordinary eager code and returns a cached machine-code binary, driving this translation through a progressive lowering architecture designed to strictly isolate high-level algorithmic decisions from low-level execution mechanics. Inheriting its organizing principle from MLIR~\cite{mlir}, Nova never translates a computation in one massive leap. Instead, the computation descends through a sequence of locally verifiable rewrites, moving from whole-tensor operations down to hardware-specific tensor-core instructions (e.g., \texttt{nova.matmul} $\rightarrow$ \texttt{linalg.matmul} $\rightarrow$ \texttt{vector.contract} $\rightarrow$ \texttt{nvgpu.mma.sync} $\rightarrow$ NVVM $\rightarrow$ PTX).

By enforcing this isolation---separating whole-graph fusion and analytic tile sizing from vectorization and tensor-core mapping---the architecture guarantees that global graph optimizations are preserved and executed exactly as intended on the physical hardware. For GPU code generation, Nova builds on the structured-codegen approach of MLIR and IREE~\cite{iree,mlircodegen}, whose tiling, vectorization, and distribution passes lower computations onto low-level instructions. However, Nova fundamentally shifts this machinery to accommodate a full, fused training step rather than a static forward inference graph.

In practice, Nova acts as an optional JIT compilation backend beneath the BluTrain deep learning framework---the layer that provides the high-level API for model definition, automatic differentiation~\cite{pytorch,tensorflow,autodiff}, neural-network modules, and optimizers. Nova does not replace this layer; the framework executes eagerly by default. When triggered, the library executes a training step, tags the resulting computation graph with its target device, and passes it to the compiler through a single, well-defined interface, which returns a compiled callable invoked in place of standard eager dispatch.

\begin{figure}[t]
  \centering
  \includegraphics[width=0.95\columnwidth]{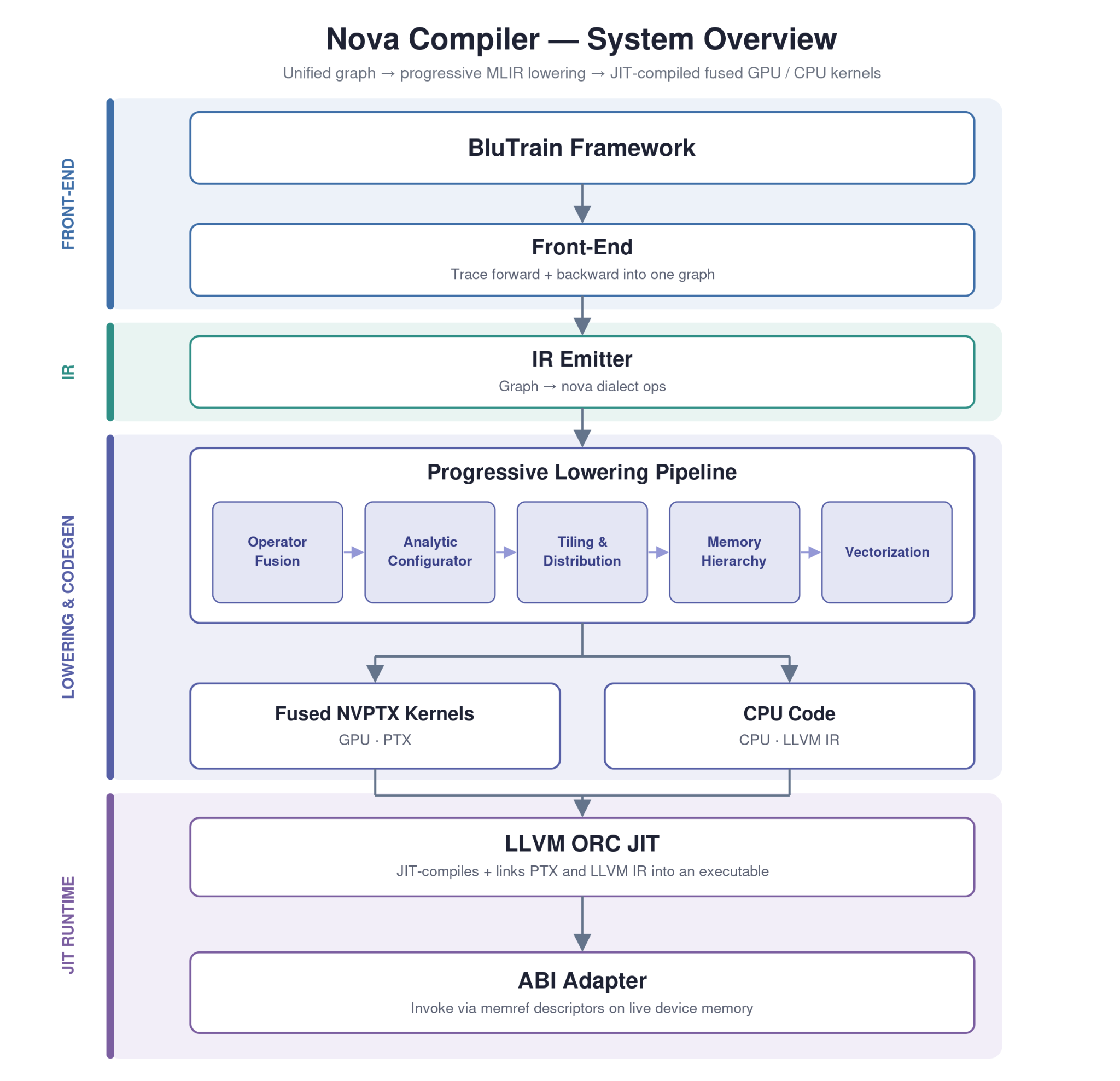}
  \caption{\textit{Nova system architecture}}
  \label{fig:arch}
\end{figure}

Figure~\ref{fig:arch} illustrates the component stack in the order a training step encounters it. This stack is cleanly delineated into four stages, each built specifically to enforce the compiler's primary mandate:

\noindent\textbf{1. Front-End Tracing:} Built to overcome the opaque dispatch of eager frameworks, the front-end dynamically traces BluTrain API calls. By capturing both forward activations and backward gradients into a unified graph, it unlocks the global visibility required for whole-step optimizations.

\noindent\textbf{2. Intermediate Representation (IR) (Section~\ref{sec:ir}):} Designed to provide a mathematically formal structure for optimization, the IR Emitter lowers the captured trace into the custom nova dialect. This represents the full training step natively within MLIR, serving as the definitive foundation for both cross-operator fusion and compiler-native gradient synchronization.

\noindent\textbf{3. Hardware Mapping and Code Generation (Section~\ref{sec:backend}):} Built to naturally orchestrate complex fusions---such as FlashAttention-style fused attention---without requiring manual kernel overrides, this pipeline enforces the compiler's absolute hardware control. An Analytic Configurator derives exact tile sizes and warp mappings directly from physical hardware limits, which are subsequently preserved through memory hierarchy passes and final vectorization to emit highly optimized NVPTX kernels. This robust pipeline also natively supports safe, opt-in mixed-precision demotion.

\noindent\textbf{4. JIT Runtime (Section~\ref{sec:runtime}):} Designed to ensure aggressive compilation does not bottleneck the dynamic training loop, the LLVM ORC JIT caches the generated artifacts. An ABI adapter binds the executable directly to live device memory, guaranteeing that the heavy cost of whole-graph optimization is paid exactly once.

\section{Graph Tracing and Intermediate Representation}\label{sec:ir}

The compilation of a deep learning model begins by converting the user's high-level training script into a format suitable for low-level optimization. This section details the frontend of the Nova pipeline: the mechanism by which disparate framework operations are ingested, combined, and represented as a single, fully differentiable computation graph within MLIR. By structurally unifying both the forward activations and backward gradient calculations into one continuous IR block, this stage establishes the algebraic foundation necessary for all subsequent backend passes.

\subsection{The nova Dialect: The MLIR Ingestion Layer}

The nova dialect was built to overcome the optimization boundaries imposed by standard eager execution, serving as the primary entry point to the Nova compiler. It provides a high-level, device-independent MLIR representation for tensor computations, enforcing strict value-semantics; it models only the pure mathematics of a training step---what to compute, not how to compute it. By defining exact algebraic building blocks (e.g., \texttt{nova.matmul}, \texttt{nova.sce}) and stripping away all hardware assumptions, the Intermediate Representation remains entirely decoupled from both framework-specific ASTs (like PyTorch's ATen) and target accelerators.

Given the existence of established MLIR dialects like TOSA or StableHLO, why construct a custom nova dialect? Standard tensor dialects are predominantly designed for static inference graphs and lack the vocabulary required to natively represent training mechanics. By defining our own dialect, we gain the extensibility to embed explicitly differentiable operations (e.g., \texttt{nova.gelu\_backward}) and compiler-driven gradient synchronization directly into the IR---capabilities that are critical for optimizing the complete, end-to-end training loop.

Crucially, because nova is natively differentiable, the frontend emits both forward activations and their derived backward gradients into the identical execution block. For example, consider the IR generated for a standard transformer language model head (Linear $\rightarrow$ GELU $\rightarrow$ Softmax Cross Entropy):

\begin{lstlisting}
Func.func @fused_training_step(%act: tensor<8x1024x384xf32>, %weight: tensor<384x50304xf32>, %bias: tensor<50304xf32>, %labels: tensor<8x1024xi16>) -> tensor<1xf32> {
// 1. Forward Pass (Linear -> GELU -> Softmax Cross Entropy)
%linear_out = nova.linear %act, %weight, %bias : tensor<8x1024x384xf32>, tensor<384x50304xf32>, tensor<50304xf32>
%gelu_out = nova.gelu %linear_out : tensor<8x1024x50304xf32>
%softmax, %loss = nova.sce %gelu_out, %labels : tensor<8x1024x50304xf32>, tensor<8x1024xi16>
// 2. Backward Pass natively woven into the same MLIR block
%loss_scale = nova.constant {value = dense<1.0> : tensor<1xf32>} : tensor<1xf32>
%grad_sce = nova.sce_backward %softmax, %labels : tensor<8x1024x50304xf32>, tensor<8x1024xi16>
%grad_scaled = nova.mul %grad_sce, %loss_scale : tensor<8x1024x50304xf32>, tensor<1xf32>
%grad_gelu = nova.gelu_backward %grad_scaled, %linear_out : tensor<8x1024x50304xf32>, tensor<8x1024x50304xf32>
%grad_input, %grad_weight, %grad_bias = nova.linear_backward %grad_gelu, %act, %weight : tensor<8x1024x50304xf32>, tensor<8x1024x384xf32>, tensor<384x50304xf32>
return %loss : tensor<1xf32>
}
\end{lstlisting}

As demonstrated in the IR snippet above, the boundary between forward computation and backpropagation is completely dissolved. By presenting the entire training step as a single function with explicitly defined data dependencies, the nova dialect provides the exact structural foundation required by the downstream compiler backend to perform aggressive whole-graph fusion.

\subsection{Automated Gradient Synchronization (Data-Parallel Integration)}\label{sec:ddp}

Fusing the entire training step into a single executable binary breaks the operational boundaries required by standard Data-Distributed Parallel (DDP) implementations. In standard eager execution, DDP achieves network-compute overlap by attaching runtime autograd hooks to individual backward operations to trigger synchronizations (e.g., NCCL all-reduces) mid-execution. Because compiling the step into a single cohesive block eliminates these eager dispatch boundaries, standard DDP would be forced to block communication until the entire backward pass completes, destroying the overlap critical for distributed scaling.

To recover this overlap without compromising the fused execution graph, gradient synchronization is embedded directly into the compiler's IR, orchestrating communication entirely at compile-time rather than relying on dynamic runtime hooks. Before lowering, the frontend statically evaluates the backward execution order and packs the parameters into contiguous communication buckets (e.g., 25\,MB).

\begin{figure}[t]
  \centering
  \includegraphics[width=\columnwidth]{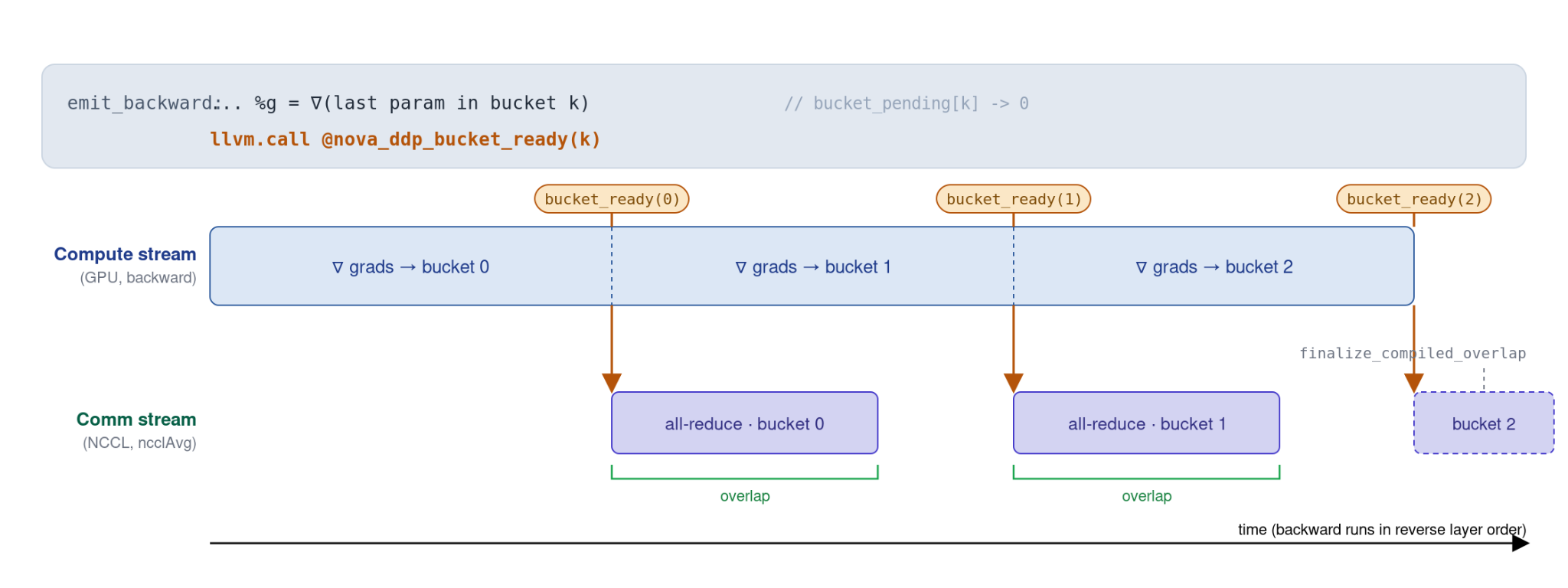}
  \caption{\textit{Compiled DDP architecture. By statically injecting an \texttt{llvm.call @nova\_ddp\_bucket\_ready(k)} the exact moment a communication bucket's final gradient is computed, Nova triggers asynchronous NCCL all-reduces to achieve perfect network-compute overlap without eager framework hooks.}}
  \label{fig:ddp}
\end{figure}

During the \texttt{emit\_backward} pass, Nova tracks the \texttt{bucket\_pending} state for each communication bucket. The exact instant the math for a bucket's final gradient is emitted, the compiler natively injects an \texttt{llvm.call} to a \texttt{nova\_ddp\_bucket\_ready} callback directly into the IR. This injected callback immediately triggers an asynchronous all-reduce across the Data Parallel replicas, as illustrated in Figure~\ref{fig:ddp}.

By enforcing synchronization via static IR-level callback injection, Nova guarantees optimal network-compute overlap and distributed scalability without sacrificing the zero-overhead performance of the JIT-compiled graph.

\section{Hardware Mapping \& Code Generation (Backend)}\label{sec:backend}

Mapping a unified tensor graph to peak hardware throughput demands architectural control over two decisions: how the computation is split into kernels, and how each of those kernels is scheduled. The Nova backend takes the first with cross-operator fusion, collapsing the training step into fewer and denser kernels, and the second with an Analytic Configurator that derives each kernel's tiling, warp mapping and memory plan from the problem shape and the queried device limits. Deriving rather than searching keeps that second decision inside a compile-time budget a JIT can pay, microseconds rather than the minutes an autotuning search would cost. This section details the systematic descent from the high-level nova dialect directly into bare-metal PTX.

\subsection{Nova to Linalg \& Operator Fusion}\label{sec:fusion}

The backend pipeline begins by lowering the \texttt{nova} dialect into linalg. This structural transition exposes the raw mathematical boundaries of the graph, enabling operator fusion across the entire training step, which the front end emits as a single MLIR function. The primary mechanism is elementwise fusion, which collapses chains of operations that share the same index space. When a producer's output feeds directly into a consumer, the compiler composes the indexing maps and inlines the mathematical body, fusing them into a single loop.

Collapsing a chain into one loop removes its intermediate tensors from the program entirely: they are never allocated, and values pass between the inlined bodies in registers. In a language-model training step the largest such intermediate is in the loss tail, where the softmax normalization and the cross-entropy gradient are separate passes over the full vocabulary logits.

\begin{figure}[t]
  \centering
  \includegraphics[width=\columnwidth]{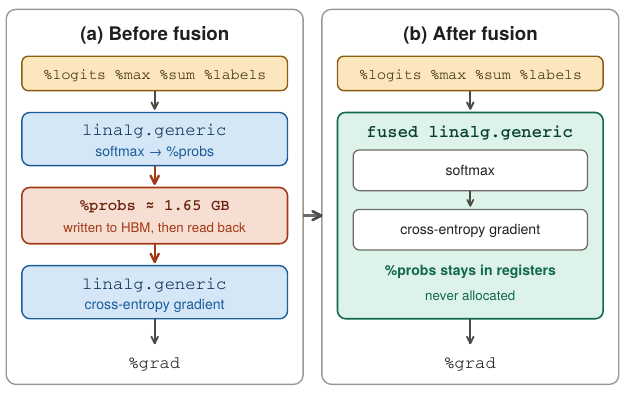}
  \caption{\textit{Elementwise fusion prevents a full-vocabulary intermediate from spilling to High Bandwidth Memory by keeping the softmax-to-gradient dependency in registers.}}
  \label{fig:fusion}
\end{figure}

For example, fusing these two operations eliminates an entire \texttt{tensor<8x1024x50304xf32>}---412 million \texttt{f32} elements, saving roughly 1.65\,GB of memory bandwidth per step. The intermediate is never materialized in memory; it remains entirely in registers, as Figure~\ref{fig:fusion} illustrates.

However, mathematical validity alone does not guarantee hardware executability. Nova's \texttt{NovaElementwiseOpFusionPass} actively analyzes indexing maps to reject fusions that the downstream GPU pipeline cannot lower correctly---such as fusing into contraction consumers or creating invalid memory access patterns for reductions.

To ensure safety and prevent redundant computation, a producer is only fused if it has exactly one use. Finally, while elementwise operations are blocked from fusing directly into a contraction's body, a dedicated pass folds residual or bias additions directly into the GEMM's destination operand. This allows a linear layer, its bias, and its residual connection to execute as a single accelerated kernel.

\noindent\textbf{The Impact of Aggressive Fusion.} The primary driver behind Nova's throughput advantages and memory optimizations (Sections~\ref{sec:microbench} and~\ref{sec:e2e}) is not faster individual kernels, but its relentless pursuit of whole-graph operator fusion. By collapsing sequences of elementwise operations, mathematical reductions, and gradient computations into single kernels, Nova systematically prevents massive intermediate tensors from ever spilling to High Bandwidth Memory (HBM). Rather than executing a training step as a sequence of isolated library calls, Nova executes it as a deeply integrated graph where outputs of one operation are held in registers to immediately feed the next.

\begin{figure}[tb]
  \centering
  \includegraphics[width=\columnwidth]{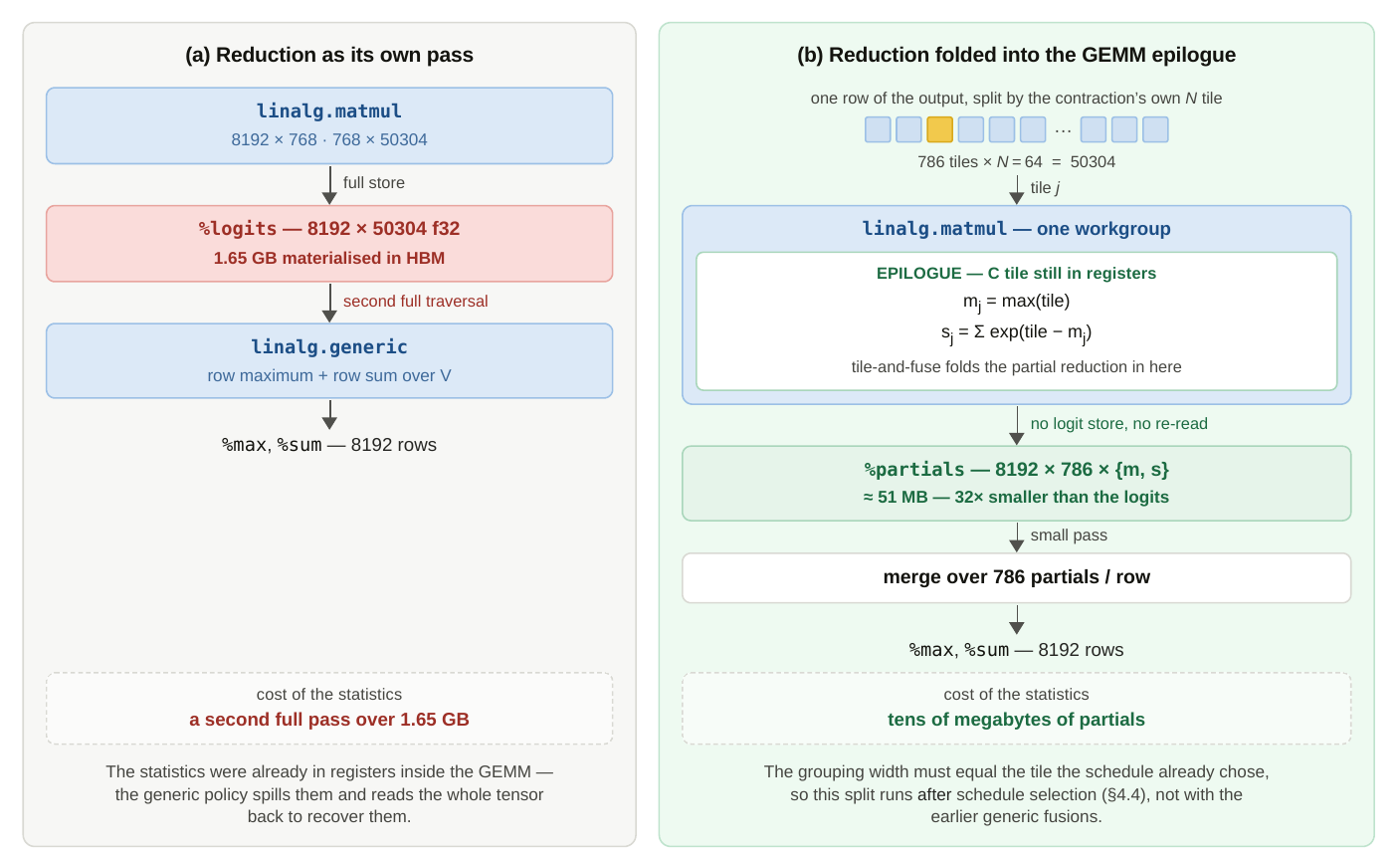}
  \caption{\textit{Fusing the loss-head reduction into the contraction epilogue. (a) As a pass of its own, the reduction forces the full 8192 $\times$ 50304 logit tensor into HBM and reads all of it back to recover statistics the GEMM already held in registers. (b) Grouped per \texttt{N} tile, the row maximum and row sum are accumulated in the GEMM's own epilogue while the C tile is still resident, leaving only 8192 $\times$ 786 partials for a small merge pass.}}
  \label{fig:epilogue}
\end{figure}

\noindent\textbf{Case Study: Fusing a Reduction into a Contraction Epilogue.} To illustrate the mechanical depth of these fusions, consider a deliberate exception to the standard fusion policy: the Softmax Cross Entropy loss head. The loss head computes logits via a massive contraction and then reduces them along the vocabulary axis to find a row maximum and sum. Because these reductions run along the identical axis the contraction produces in tiles, they do not require a second memory traversal. Instead, each workgroup accumulates the statistics for the \texttt{N}-tile it has just computed directly inside the GEMM's own epilogue. Nova implements this split as a dedicated pass. An exponential reduction over a contraction is grouped into per-tile partial statistics, which the tile-and-fuse mechanism folds into the GEMM kernel, followed by a small merge over the partials. Because this grouping must strictly match the contraction's chosen \texttt{N} tile, the split executes after schedule selection (Section~\ref{sec:config}) rather than alongside earlier, generic fusions, and the resulting merge operations are configured by a second pass over the IR. On the Ada 6000 configuration measured in Section~\ref{sec:eval}, the partials are 8192 $\times$ 786 rather than the full 8192 $\times$ 50304 logits. As a result, computing the statistics adds only tens of megabytes in intermediate storage instead of forcing a second full memory pass over 1.65\,GB. Figure~\ref{fig:epilogue} contrasts the two paths.

Crucially, this merge operation cannot be generic. A running maximum and a running sum do not combine via the same operation that accumulates them: before two partial sums can be added, one must be rescaled by the exponential of the difference of their maxima. Standard partial-reduction support assumes the merge is structurally identical to the fold, which would produce silently incorrect gradients in this context. To prevent this, the pass supplies a custom, mathematically-aware merge region. Attention is handled in the same spirit, relying on a dedicated lowering rather than the generic policy (Section~\ref{sec:attention}).

\subsection{Attention as a First-Class Lowering}\label{sec:attention}

Elementwise fusion (as discussed in Section~\ref{sec:fusion}) can't handle attention because the softmax operation sits right between two matrix contractions, and the fusion policy strictly avoids merging across a contraction in either direction. Nova therefore lowers attention through a dedicated path, and that path has to run before the standard Nova-to-Linalg conversions: those conversions mark the entire \texttt{nova} dialect illegal and register no pattern for \texttt{nova.attention}, so the operation must be rewritten first. Rather than falling back to an external library call, the path generates a standard tiled loop nest that the rest of the compiler optimizes just like any other code.

\noindent\textbf{Building the Online Softmax.} The first step transforms \texttt{nova.attention} into a lower-level \texttt{nova.online\_attention} operator. This operator uses indexing maps to handle the data layouts for the Query, Key and Value tensors. Rather than running a separate transpose operation, the compiler changes how it reads the data, making one dimension parallel and another a reduction. The transpose comes for free. The same mechanism absorbs the head-major repacking step that normally precedes attention, eliminating three transpose kernels per layer.

The operation sets up three running accumulators: the output and the sum, both starting at zero, and the maximum, starting at a large finite negative number rather than negative infinity. For causal attention it builds a rank-2 tensor of 8-bit keep-flags; because the mask is rank-2, one copy broadcasts across every batch and head rather than being replicated per head. The operation also saves the log-sum-exp of each row for the backward pass to reuse.

When the compiler tiles the key sequence dimension ($K_2$), it breaks the operation down into ten \texttt{linalg.generic} operations per tile. Two adjustments to the textbook formulation matter here:

\begin{itemize}
  \item \textbf{Base-2 softmax.} Softmax is computed in base 2 instead of base $e$. The conversion factor is folded into the initial scale, so no conversion appears inside the hot loop, and the result converts back to base $e$ only once at the very end.
  \item \textbf{Masking.} Masked tokens are given a large finite negative bias ($-10^{30}$) rather than negative infinity. These kernels are compiled under fast-math (\texttt{ninf}) semantics, which makes a true $-\infty$ undefined; the finite sentinel underflows cleanly to zero during the exponentiation step.
\end{itemize}

\begin{lstlisting}
// One K2 tile of the fused forward pass.
// Shapes from a GPT-2 step: wgM = 64, K2 tile = 32, head dim = 64.
scf.for %k2 = %c0 to %ub step %c32
    iter_args(%acc = %acc0, %mx = %mx0, %sm = %sm0) {
  // 1. QK contraction, scaled by (scale * log2(e))
  %s = linalg.generic { ... } ins(%q, %k) outs(%s0)
         attrs = {nova.causal, nova.flash_mma}
  // 2. Apply causal mask (adds -1.0e30 to masked lanes)
  %sm4 = linalg.generic ins(%m) outs(%s) { ... }
  // 3. Update running maximum
  %mx1 = linalg.generic { /* m_new = max(S, m_old) */ }
  // 4. Calculate normalization factor
  %norm = linalg.generic { /* exp2(m_old - m_new) */ }
  // 5. Scale the running sum
  %sm1 = linalg.generic { /* sum <- norm * sum */ }
  // 6. Exponentiate scores in-place
  %p = linalg.generic ins(%mx1) outs(%sm4) { /* P = exp2(S - m_new) */ }
  // 7. Add to running sum
  %sm2 = linalg.generic { /* sum <- sum + rowsum(P) */ }
  // 8. Scale previous output
  %acc1 = linalg.generic { /* O <- norm * O */ }
  // 9. PV contraction
  %acc2 = linalg.generic { ... } ins(%p, %v) outs(%acc1) attrs = {nova.flash_mma}
  scf.yield %acc2, %mx1, %sm2
} {nova.causal_clamped}
\end{lstlisting}

\noindent\textbf{Keeping Scores in Registers.} Simply breaking the operation into tiles isn't enough to keep the data out of memory; the compiler also has to control exactly how those tiles are distributed across the GPU threads.

Both attention contractions are tagged so that the subgroup tiling level does not split them into a warp-level loop. The warp split is carried in the tensor layout instead, taken from each operation's own subgroup tile. As a result the entire chain (QK contraction, softmax, PV contraction) stays inside a single block-level region, and the intermediate score tile ($S$) and probability tile ($P$) live entirely in registers for their whole lifetime, never written to shared or global memory. The final division $O = (1/\mathrm{sum}) \cdot O$ is not part of the decomposition at all: it is emitted by the reduction merge and then sunk into the kernel's epilogue, which removes the extra kilobyte of shared memory it would otherwise have required.

\noindent\textbf{Exploiting Causality Twice More.} The compiler treats the causal mask as a structural rule rather than as data, in two independent rewrites. Both produce results bit-identical to the unmodified form.

\begin{figure}[tb]
  \centering
  \includegraphics[width=\columnwidth]{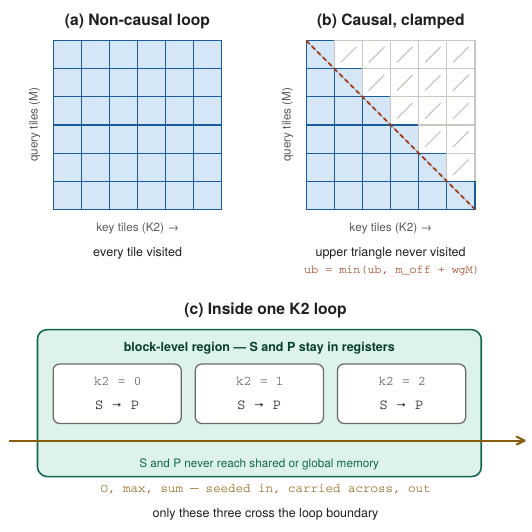}
  \caption{\textit{Causal clamping and register residency. (a) A non-causal loop visits every key tile. (b) Clamping the upper bound to \texttt{min(ub, m\_off + wgM)} eliminates the tiles above the diagonal rather than predicating them; tile shapes are schematic, while the selected configuration pairs a 64-row query tile with a 32-wide $K_2$ step. (c) Within one $K_2$ loop the score tile $S$ and probability tile $P$ are confined to a single block-level region, so only $O$, the running maximum and the running sum cross the loop boundary.}}
  \label{fig:attention}
\end{figure}

\begin{itemize}
  \item \textbf{Eliminating the mask read.} Instead of reading the mask from memory, the compiler recomputes the mask boundary on the fly from the loop indices: the tile's slice offset plus \texttt{linalg.index}. Dropping the read leaves the loop that built the mask with no consumers, so it is deleted and the mask tensor is never materialized.
  \item \textbf{Shortening the loop.} In causal attention everything above the diagonal is masked out. Rather than computing those values and throwing them away, the compiler clamps the loop's upper bound so those tiles are never visited (Figure~\ref{fig:attention}); in the backward direction it raises the lower bound instead. Asymptotically this halves the score and value contractions. The work is eliminated rather than predicated.
\end{itemize}

\noindent\textbf{The Backward Pass.} The backward pass is also entirely fused. It compiles down into two kernels: one computes the Query gradients, reducing over keys, and the other computes the Key and Value gradients, reducing over queries. Neither kernel stores or reads the probability matrix from memory. Each recomputes it inside the kernel from the log-sum-exp cached by the forward pass, so the score matrix never exists in memory in either direction.

\noindent\textbf{Configuring the Hardware Schedule.} For standard contractions the Analytic Configurator (discussed in Section~\ref{sec:config}) picks tile sizes based on arithmetic intensity. Attention is the one exception. Its tiles are coupled, since the score tile, both statistics vectors and the output accumulator all live in the same block-level region, so the binding constraint is the shared memory consumed by the whole fused loop rather than either contraction's intensity. With $Q$, $K$ and $V$ staged in shared memory that footprint is $(\mathrm{wgM} + 2 \cdot K_2\mathrm{tile}) \cdot d \cdot 4$ bytes for head dimension $d$. The configurator therefore overrides the intensity-derived shape with a smaller one, a fixed 64-row query tile and a 32-wide $K_2$ step, from which the shared $K_2$ extent and the subgroup split are recomputed. The smaller footprint leaves room for more resident thread blocks per SM on a given target, at the cost of less work per block. The shape is fixed rather than solved for, pending a budget model that accounts for the driver's per-block reserve.

\subsection{Mixed Precision}

To extract the theoretical throughput gains of reduced-precision (bfloat16) tensor cores without destabilizing the numerical integrity of the training step, mixed precision is implemented as a targeted, opt-in IR pass executed before the Analytic Configurator. The pass systematically demotes only the input operands of matrix contractions to bfloat16, leaving all accumulators, elementwise activations, and normalizations in full precision (float32).

Because this transformation occurs early in the pipeline, the Analytic Configurator naturally observes the halved byte-width of the operands. It responds by selecting a bfloat16 \texttt{mma.sync} intrinsic and a larger tile shape, since the shared-memory footprint per tile is halved. The low-precision tile seeds are tuned separately from the float32 path.

\subsection{The Analytic Configurator (Strategy \& Tiling)}\label{sec:config}

To avoid the compilation overhead and architectural fragility of runtime autotuning, Nova relies on an Analytic Configurator. A contraction's schedule follows from its shape, its element type and the target's published limits, so every candidate is scored by a closed-form model at compile time rather than measured by trial execution. Before any lowering occurs, the configurator reads the queried device parameters: SM count, static and dynamic shared-memory capacity, shared memory and registers per SM, resident warp and block limits, L2 capacity, clock rate, DRAM bandwidth, and the available tensor-core shapes. It then evaluates each contraction in the IR and classifies the workload by its Arithmetic Intensity (AI):
\[
\mathrm{AI} = \frac{2 \cdot M \cdot N \cdot K}{M \cdot K + K \cdot N + M \cdot N}
\]

Contractions with low AI are bottlenecked by memory bandwidth and receive smaller tile shapes to maximize occupancy. High AI workloads are compute-bound and receive larger tiles to fully saturate the tensor cores. The configurator selects tile sizes, warp distributions and MMA intrinsics deterministically, embedding the resulting schedule directly into the IR as a \texttt{lowering\_config} attribute. The attention contractions are the one exception, scheduled from bounded seeds against the shared-memory budget of their fused loop instead (Section~\ref{sec:attention}).

On the independent M/N distribution path, the chosen tile is then scored against the device rather than accepted outright. Nova divides peak tensor-core throughput by DRAM bandwidth to obtain the machine's balance point, and compares that against the schedule's own operational intensity. The traffic side of that ratio is tile-aware rather than shape-only. An operand panel small enough to stay resident in L2 is counted once; a panel that exceeds L2 is counted once per pass the tiling makes over it. When a schedule falls on the memory side of the balance point, the configurator raises residency by distributing the same tile across more warps instead of shrinking it. Tile size determines operand reuse, so trading it away for occupancy costs more in re-fetched operands than the extra resident warps recover.

For instance, when the Configurator evaluates a $4096 \times 4096 \times 4096$ matrix multiplication (AI $\approx$ 2730) on the RTX 6000 Ada, it classifies the workload as highly compute-bound and assigns $256 \times 128$ block tiles to saturate the SMs. Because this tile scores comfortably above the machine's balance point, no occupancy refinement is needed. The following snippet shows the complete hardware schedule, from the thread block grid down to the shared-memory promotion strategy, preserved natively within the IR:

\begin{lstlisting}
%C = linalg.matmul {lowering_config = {
  workgroup = [256, 128, 0],   // each thread block computes a 256x128 output tile
  reduction = [0, 0, 64],      // march over K in steps of 64
  subgroup = [64, 64, 0],      // 64x64 of that tile per warp
  wg_subgroup = [4, 2, 0],     // distributed over a 4x2 grid of warps
  mma_kind = 5 : i32,          // MMA_SYNC_TF32_16x8x8
  promoted_operands = [0, 1],  // stage A and B in shared memory
  pad_operands, thread = [0, 0, 0]
}} ins(%A, %B) outs(%C0) -> tensor<4096x4096xf32>
\end{lstlisting}

\subsection{Vectorization \& Bufferization}

To bridge the gap between high-level loops and the strict layouts required by NVIDIA Tensor Cores, Nova rewrites scalar loops directly into the vector dialect, completely avoiding reliance on closed-source backend libraries.

Because Tensor Cores require operands in rigid, interwoven layouts distributed precisely across a warp's 32 lanes, a dedicated layout pass injects explicit \texttt{\#nova\_vector\_ext.nested\_layout} attributes. This mathematically shatters logical tiles into hardware registers at compile time, avoiding runtime index arithmetic and allowing the compiler to explicitly cast memory-bound tensors into distributed layouts before contraction, as demonstrated in the resulting vector IR snippet:

\begin{lstlisting}
// Explicitly casting logical tiles to distributed hardware layouts
%laid_out_A = nova_vector_ext.to_layout %slice_A to layout(#nested) : vector<64x32xf32>
%laid_out_B = nova_vector_ext.to_layout %slice_B to layout(#nested1) : vector<32x32xf32>
// 2. The scalar loop is directly rewritten into a hardware-mapped vector contraction
%c = vector.contract {
indexing_maps = [#mapA, #mapB, #mapC],
iterator_types = ["parallel", "parallel", "reduction"],
kind = #vector.kind<add>
} %laid_out_A, %laid_out_B, %acc : vector<64x32xf32>, vector<32x32xf32> into vector<64x32xf32>
\end{lstlisting}

Bufferization then assigns memory spaces by looking at how each value is consumed: a buffer feeding a warp- or thread-mapped loop inside a block-mapped one becomes a shared-memory allocation, materializing the promotion the Configurator already recorded, while everything else stays unannotated and lowers to registers. Lane-level distribution runs after bufferization, turning each laid-out tile into the per-thread fragments the MMA instructions consume.

\subsection{Register-Level Optimizations \& Pipelining}

\begin{figure}[!tb]
  \centering
  \includegraphics[width=\columnwidth]{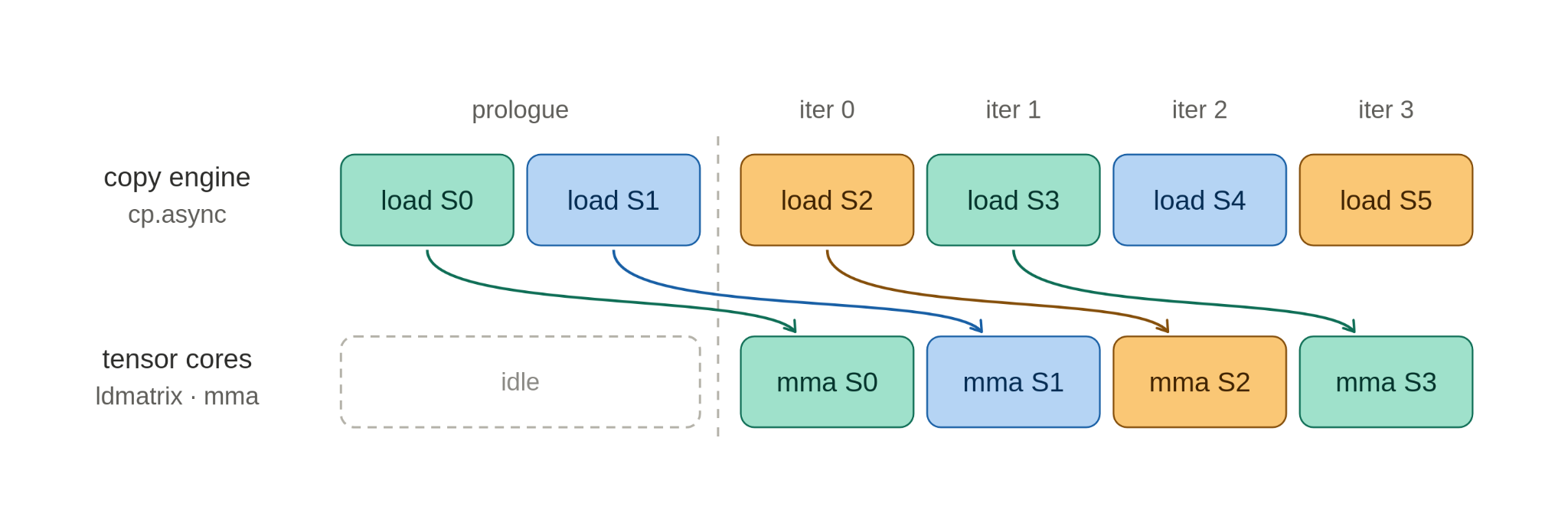}
  \caption{\textit{Software-pipelined K-loop execution. By maintaining depth $-$ 1 asynchronous copy groups in flight, the tensor cores compute on resident slab $i$ while the memory controller simultaneously prefetches future slabs, effectively hiding HBM latency.}}
  \label{fig:pipeline}
\end{figure}

Raw tensor core invocation is mathematically sufficient, but practically insufficient if the cores are starved for data. Native MLIR lowering produces mathematically correct kernels, but without explicit hardware management, they cannot compete with hand-tuned libraries like cuBLAS. To bridge this gap, Nova injects highly specialized, hardware-aware optimizations directly into its backend pipeline:

\noindent\textbf{1. Adaptive Multi-Buffering \& Software Pipelining:} The central K-loop is software-pipelined to maintain continuous data flow. Nova time-shifts \texttt{nvgpu.device\_async\_copy} operations by depth $-$ 1 iterations, guaranteeing that iteration $i{+}2$ prefetches from HBM while iteration $i$ computes on resident memory. Crucially, this depth adaptively scales; if the footprint exceeds the per-block shared memory cap, the pipeline seamlessly falls back from 3 stages to 2. Figure~\ref{fig:pipeline} shows the resulting schedule, and the structure is natively represented in the generated IR:

\begin{lstlisting}
//Pipelined K-Loop with async copies and wait groups
scf.for %k = 0 to %bound step 32 {
nvgpu.device_async_copy %B[...], %sharedB[%next_slot, ...] // prefetch future slab
nvgpu.device_async_create_group
%fr = nvgpu.ldmatrix %sharedB[%cur_slot, ...] // read resident slab
%acc = nvgpu.mma.sync(%aFrag, %fr, %acc) // tensor cores work...
nvgpu.device_async_wait { numGroups = 2 } // ...while loads are in flight
}
\end{lstlisting}

% Fig. 7 is declared ahead of the swizzling paragraph so it lands on p8
% with Fig. 6, beside the discussion that references it.
\begin{figure}[!b]
  \centering
  \includegraphics[width=0.95\columnwidth]{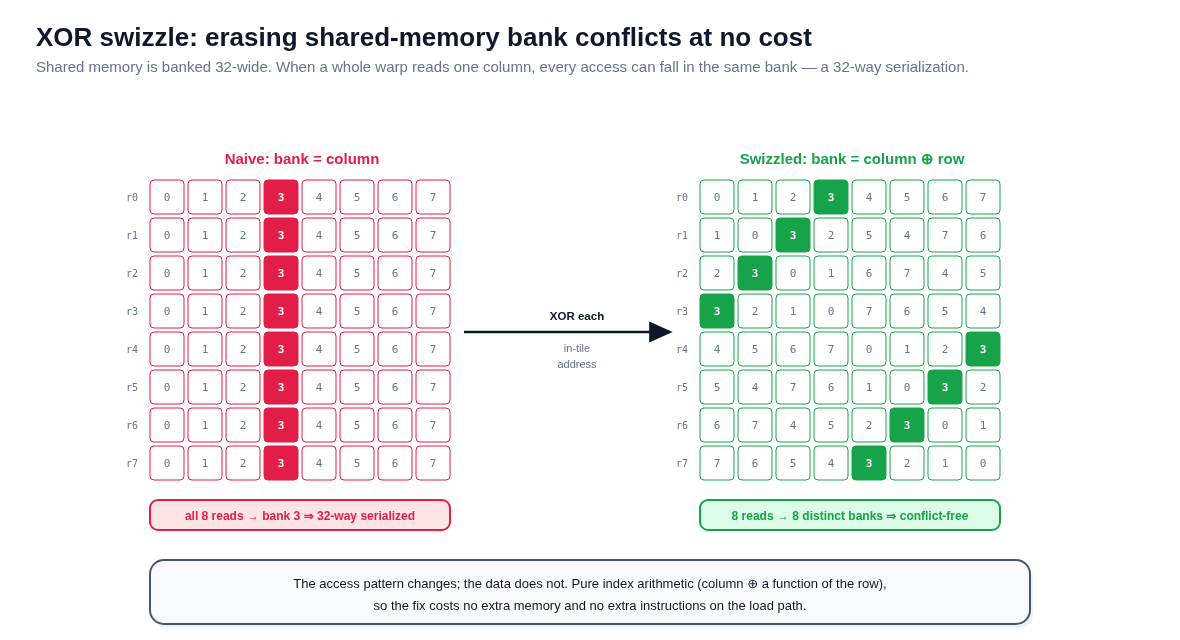}
  \caption{\textit{A warp reading one tile column. Left: with bank $=$ column, all rows land in the same bank --- serialized. Right: with the XOR swizzle bank $=$ column $\oplus$ row, the column scatters across distinct banks --- conflict-free.}}
  \label{fig:swizzle}
\end{figure}

\noindent\textbf{2. Shared Memory Swizzling:} A tensor core's throughput relies heavily on feeding it data from shared memory as quickly as possible via \texttt{ldmatrix}. However, naive column-major reads by a warp cause severe 32-way memory bank conflicts because multiple threads attempt to access the same physical memory bank simultaneously, forcing the hardware to serialize the load. To prevent this, \texttt{NovaGPUSwizzleSharedMemoryPass} injects an XOR permutation into the in-tile access logic: bank $=$ column $\oplus$ ((row \& mask) $\ll$ shift). This mathematical swizzle guarantees that elements residing in the same logical column are physically scattered across distinct hardware banks, as Figure~\ref{fig:swizzle} shows. By folding this arithmetic directly into the memory indices at compile time, Nova achieves conflict-free shared memory bandwidth with strictly zero layout overhead. This swizzle logic is realized at compile time via the following IR pattern:

\begin{lstlisting}
//XOR swizzle folded directly into shared memory indices
%c7 = arith.constant 7 : index
%c2 = arith.constant 2 : index
%masked = arith.andi %row, %c7 : index
%shifted = arith.shli %masked, %c2 : index
%swizzled_col = arith.xori %col, %shifted : index
%fr = nvgpu.ldmatrix %sharedB[%row, %swizzled_col] {numTiles = 4 : i32, transpose = false} : memref<128x32xf32, 3> -> vector<4x1xf32>
\end{lstlisting}

\noindent\textbf{3. Precision-Aware ldmatrix Routing:} When loading fragments from shared memory to the tensor cores, Nova dynamically alters its instruction emission based on operand precision and tensor layout. The hardware \texttt{ldmatrix} instruction is highly rigid, and its \texttt{ldmatrix.trans} variant is strictly limited to 16-bit granularity. Because full-precision float32 (TF32) operands would trigger a hardware fault if passed to \texttt{ldmatrix.trans}, Nova's layout distribution pass analytically routes the shared-memory reads between accelerated \texttt{ldmatrix} intrinsics and fallback scalar loads depending on the exact matrix layout (e.g., NN, NT, TN). Table~\ref{tab:ldmatrix-routing} summarises the routing.

\begin{table}[H]
\centering
\footnotesize
\setlength{\tabcolsep}{4pt}
\begin{tabular}{@{}lll@{}}
\toprule
\textbf{Contraction} & \textbf{F32 (TF32)} & \textbf{BF16} \\
\textbf{Layout} & \textbf{Emission} & \textbf{Emission} \\
\midrule
NN (Row-Major, & A: \texttt{ldmatrix} & A: \texttt{ldmatrix} \\
Col-Major) & B: Scalar Loads & B: \texttt{ldmatrix.trans} \\
\addlinespace
NT (Row-Major, & A: \texttt{ldmatrix} & A: \texttt{ldmatrix} \\
Row-Major) & B: \texttt{ldmatrix} & B: \texttt{ldmatrix} \\
\addlinespace
TN (Col-Major, & A: Scalar Loads & A: \texttt{ldmatrix.trans} \\
Col-Major) & B: Scalar Loads & B: \texttt{ldmatrix.trans} \\
\bottomrule
\end{tabular}
\caption{\textit{Precision-aware \texttt{ldmatrix} emission selected by contraction layout and operand precision}}
\label{tab:ldmatrix-routing}
\end{table}

By embedding this routing logic directly into the compiler's backend, Nova guarantees hardware compliance across all mixed-precision and full-precision paths without requiring the developer to manually manage memory constraints. This routing is injected during lowering, as shown in this MLIR snippet:

\begin{lstlisting}
//bfloat16 mixed-precision path accelerating NT layout via ldmatrix.trans
%bf16_frag = nvgpu.ldmatrix %sharedB[%row, %col] {numTiles = 4 : i32, transpose = true} : memref<128x32xbf16, 3> -> vector<4x2xbf16>
\end{lstlisting}

\noindent\textbf{4. Butterfly Warp Reductions:} Reductions bypass the MMA pipeline and lack a native tensor-core intrinsic. To optimize them, Nova pins the accumulators directly in registers and expands clustered subgroup reductions into \texttt{gpu.shuffle} butterfly trees. This maps the reduction directly onto native PTX warp-level instructions, collapsing the row entirely in registers without traversing shared memory. The expansion into native warp instructions is captured in the backend IR as follows:

\begin{lstlisting}
//Clustered subgroup reduction expanded into PTX warp shuffles
%c16 = arith.constant 16 : i32
%c32 = arith.constant 32 : i32
%shfl, %valid = gpu.shuffle xor %acc, %c16, %c32 : f32
%acc_next = arith.addf %acc, %shfl : f32
\end{lstlisting}

\section{Runtime Binding}\label{sec:runtime}

Nova isolates its heavy compilation pipeline from the hot execution path using a zero-overhead structural runtime. By guaranteeing that code generation occurs exactly once, the runtime ensures that GPU throughput gains are never bottlenecked by CPU dispatch latency during the steady-state training loop.

Deep learning graphs are highly dynamic in their memory allocations, with raw pointers cycling constantly between forward and backward passes. A naive cache keyed on memory addresses would suffer continuous cache misses, triggering catastrophic recompilations on every step.

\subsection{Structural Hashing for JIT Caching}

To solve this, Nova introduces an address-independent Structural Hashing mechanism. When a captured graph enters the runtime, the hasher explicitly ignores all concrete tensor pointers. Instead, it computes a hash based entirely on graph topology (opcodes, normalized IDs, sorted attributes) and static metadata (tensor shapes, strides, dtypes, and target devices).

Because a model's architecture remains static while its memory shifts, this structural hash guarantees a 100\% cache hit rate after the first iteration. The expensive, whole-graph compilation pipeline (detailed in Sections~\ref{sec:ir} and~\ref{sec:backend}) executes exactly once; every subsequent iteration simply retrieves the LLVM ORC JIT cached executable in microseconds.

\subsection{Graduated Argument Binding}

Even when the JIT executable is successfully retrieved from the cache, the runtime must still supply the compiled kernel with an updated memref descriptor for every active tensor. Reconstructing and repacking hundreds of tensor descriptors on the CPU side during every dispatch would reintroduce a severe synchronization bottleneck.

\section{System Evaluation}\label{sec:eval}

We evaluate Nova across two tiers: standalone kernel microbenchmarks and end-to-end data-parallel training. All evaluations were conducted on NVIDIA RTX 6000 Ada GPUs, using a single card for the kernel benchmarks and eight cards for end-to-end training.

\subsection{Kernel Microbenchmarks}\label{sec:microbench}

A transformer training step decomposes into three kinds of kernel: matrix contractions, bandwidth-bound elementwise and reduction operations, and attention. We evaluate each against the hand-written implementation a framework would otherwise dispatch to for it, and report each in the metric that governs it: throughput for contractions, achieved bandwidth for the memory-bound operators, and latency for attention. All measurements follow a warmup that excludes JIT compilation.

\noindent\textbf{1. Matrix Multiplication.}

We measure TF32 throughput on nine shapes: three squares as hardware characterisation, and six contractions drawn from the GPT-2 124M step at $d = 768$. These cover the two MLP projections and the language-model head in its forward and data-gradient orientations, with the output projection and the head's data gradient each measured at a second, smaller batch size. Baselines are cuBLAS, reached through \texttt{torch.compile}, and a Triton kernel generated by TorchInductor~\cite{pytorch2}. Exact dimensions for every case are listed in Table~\ref{tab:matmul}.

Figure~\ref{fig:matmul} reports throughput for all nine shapes. Overall, Nova achieves performance parity with both hand-tuned baselines, but its distribution reveals a distinct alignment with the high-aspect-ratio workloads that dominate large-batch LLM training. At the larger of the two measured batch sizes ($M = 16384$), Nova outperforms both cuBLAS and Triton across the board. On the MLP projections it reaches up to 80.1 TFLOPS (against 67.1 for cuBLAS and 67.8 for Triton). It similarly leads both baselines on the language-model head in the forward pass (68.7, against 65.4 for the stronger baseline) and in the data-gradient pass (74.4, against 70.6).

% matmul is Fig. 8; the bandwidth/latency/attention floats (Figs. 9-11) are
% declared earlier for placement, so the counter is set here and restored.
\setcounter{figure}{7}
\begin{figure}[H]
  \centering
  \includegraphics[width=\columnwidth]{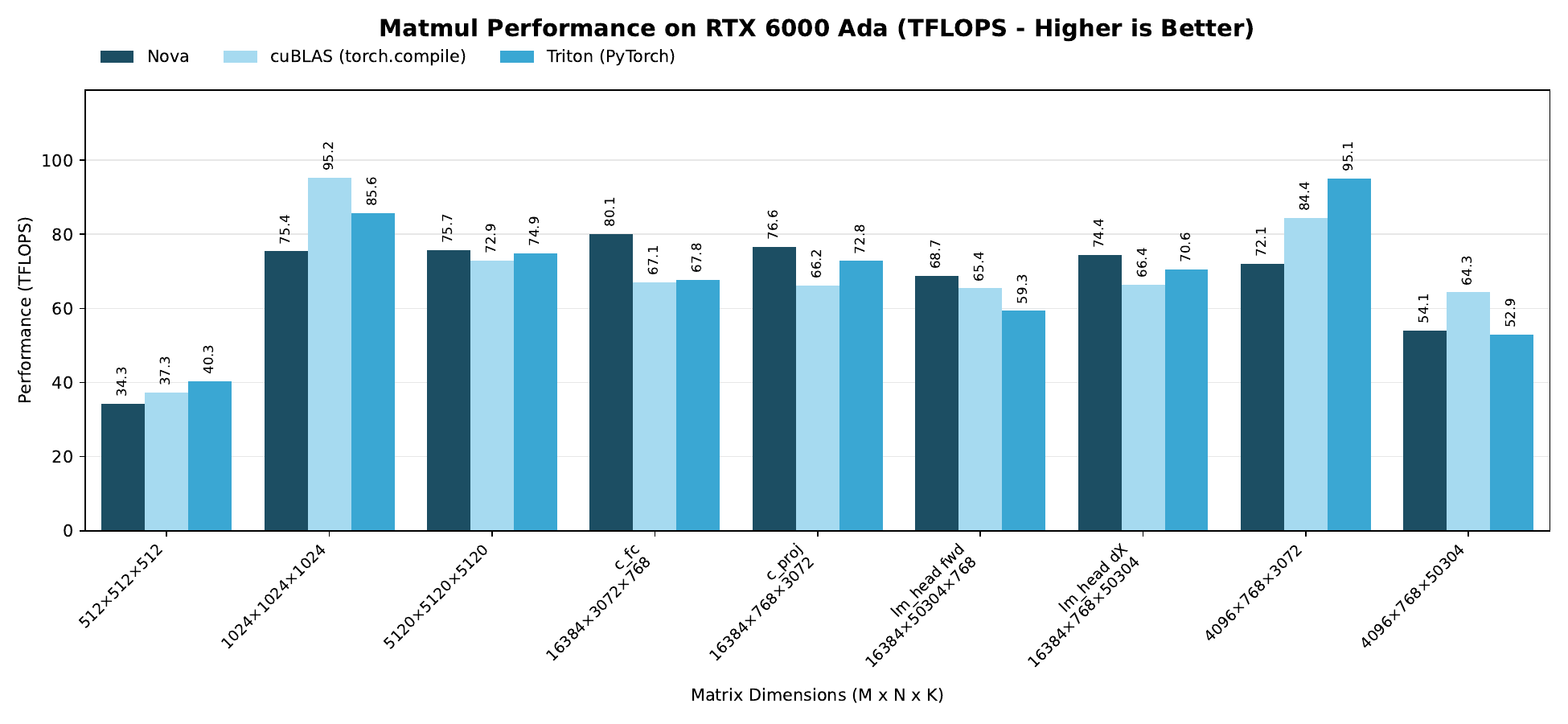}
  \caption{\textit{Standalone TF32 matrix-multiplication throughput on a single RTX 6000 Ada, comparing Nova against cuBLAS (via \texttt{torch.compile}) and a Triton kernel generated by TorchInductor. Three square shapes characterise the device; the remaining six are contractions taken from the GPT-2 124M training step.}}
  \label{fig:matmul}
\end{figure}
\setcounter{figure}{11}

% Table 2 is pinned directly beneath the matmul figure so the error table
% sits with the throughput chart it accompanies.
\begin{table}[H]
\centering
\footnotesize
\setlength{\tabcolsep}{4pt}
\begin{tabular}{@{}lccc@{}}
\toprule
\textbf{M$\times$N$\times$K} & \textbf{Nova} & \textbf{CuBLAS} & \textbf{Triton} \\
 & & \textbf{(Torch)} & \textbf{(via PyTorch)} \\
\midrule
$512^3$    & 1.88e-4 & 2.41e-4 & 6.00e-4 \\
$1024^3$   & 2.85e-4 & 2.33e-4 & 8.16e-4 \\
$5120^3$   & 3.02e-4 & 3.40e-4 & 7.05e-4 \\
\addlinespace
16384$\times$3072$\times$768   & 1.78e-4 & 2.09e-4 & 7.84e-4 \\
16384$\times$768$\times$3072   & 3.03e-4 & 2.54e-4 & 7.48e-4 \\
16384$\times$50304$\times$768  & 3.10e-4 & 2.26e-4 & 9.18e-4 \\
\addlinespace
16384$\times$768$\times$50304  & 2.92e-4 & 2.81e-4 & 8.94e-4 \\
4096$\times$768$\times$50304   & 4.12e-4 & 2.15e-4 & 7.25e-4 \\
4096$\times$768$\times$3072    & 3.12e-4 & 2.42e-4 & 7.13e-4 \\
\bottomrule
\end{tabular}
\caption{\textit{Maximum relative error against a double-precision reference, per shape and implementation.}}
\label{tab:matmul}
\end{table}

The deficits fall on smaller problem sizes, specifically the smaller square shapes and the batch-4 contractions. At $1024^3$, Nova achieves 75.4 TFLOPS against cuBLAS's 95.2. On the batch-4 MLP projection ($4096 \times 768 \times 3072$), it trails Triton by 24\% (72.1 against 95.1). However, as shapes scale up, Nova's analytic schedule recovers its competitiveness even on square workloads, slightly edging out cuBLAS and Triton at $5120^3$ (75.7 TFLOPS against 72.9 and 74.9). The two measured batch sizes bracket the configuration we train, which runs at $B = 8$ and so presents $M = 8192$ to every contraction, above the size at which Nova's schedule takes the lead. The gap at the smaller size is therefore the least consequential place to lose: the large asymmetric contractions are what dictate end-to-end training throughput.

Correctness was verified against a double-precision reference: for each shape, 64 sampled output elements are recomputed as fp64 dot products on the host and the maximum relative error reported. As Table~\ref{tab:matmul} shows, all three implementations remain well inside the $2 \times 10^{-3}$ tolerance implied by TF32's ten mantissa bits. Nova's error is lower than the Triton kernel's on every shape, by factors of 1.8 to 4.4, and lower than cuBLAS's on three of the nine; its worst case is $4.1 \times 10^{-4}$, against a table-wide worst of $9.2 \times 10^{-4}$ which belongs to Triton.

% Benchmark double-column floats are declared here (ahead of the memory-bound
% subsection) so the two-column output routine releases them onto pages 11-12,
% beside the discussion that references them.
% Their numbers are set explicitly: matmul is Fig. 8 (declared in the
% microbenchmark subsection), these are Figs. 9-11, and the counter is
% restored afterwards so the end-to-end figures continue at 12.
\setcounter{figure}{8}
\begin{figure*}[!t]
  \centering
  \includegraphics[width=0.325\textwidth]{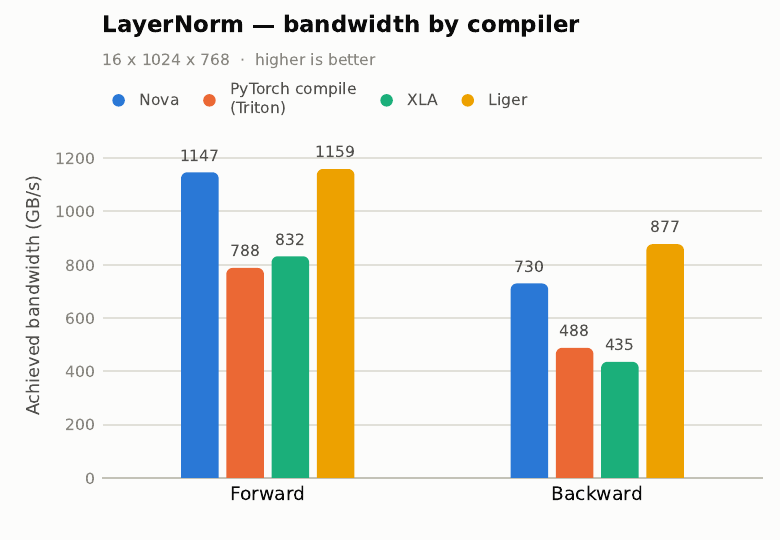}\hfill
  \includegraphics[width=0.325\textwidth]{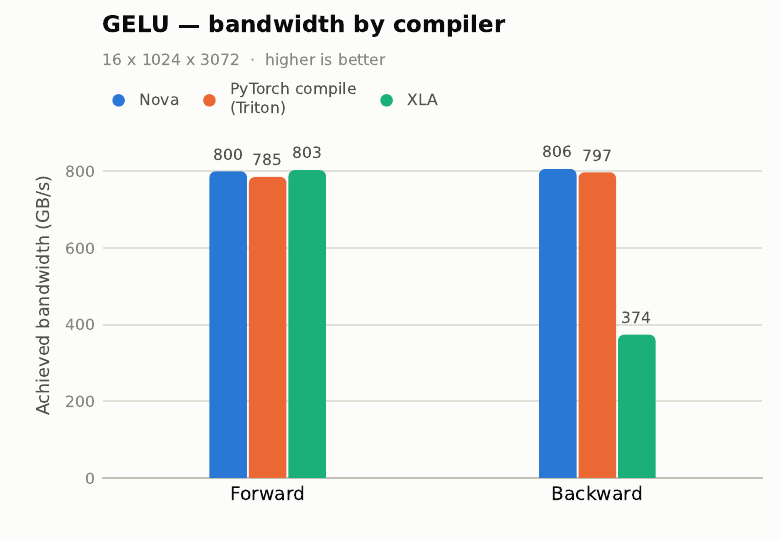}\hfill
  \includegraphics[width=0.325\textwidth]{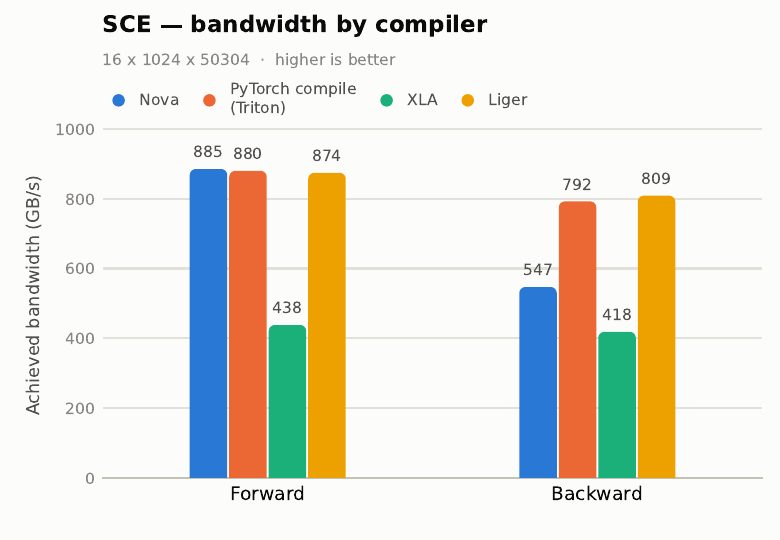}
  \caption{\textit{Achieved bandwidth for the memory-bound operators, forward and backward; higher is better. Liger provides no GELU kernel. LayerNorm's forward figures exceed the card's 960\,GB/s DRAM peak because its 50\,MB activation is largely L2-resident.}}
  \label{fig:membw}
\end{figure*}

\begin{figure*}[!b]
  \centering
  \includegraphics[width=0.325\textwidth]{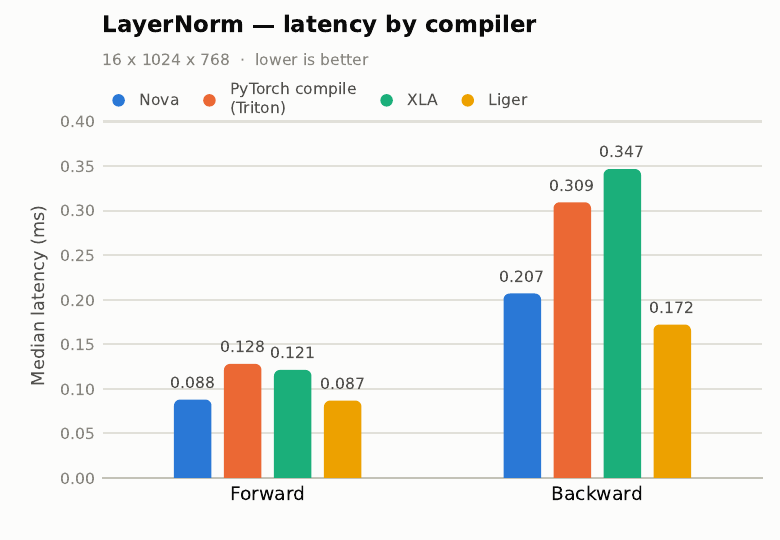}\hfill
  \includegraphics[width=0.325\textwidth]{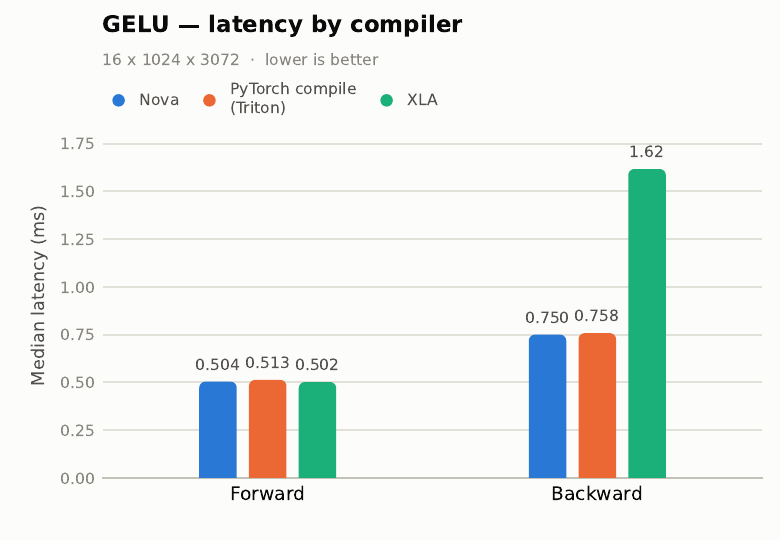}\hfill
  \includegraphics[width=0.325\textwidth]{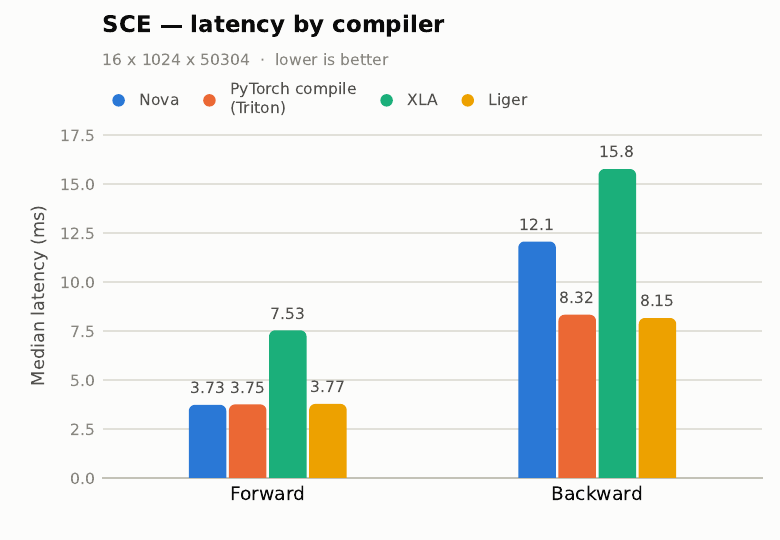}
  \caption{\textit{Kernel latency for the same operators. Note the scale difference between panels: the cross-entropy pair costs two orders of magnitude more than LayerNorm, so it dominates a training step even where per-byte efficiency is comparable.}}
  \label{fig:memlat}
\end{figure*}

% memlat is Fig. 10; Fig. 11 (attention) is declared later, at the Attention
% subsection, so the counter stops at 10 here rather than 11.
\setcounter{figure}{10}

To eliminate this overhead, Nova employs Graduated Argument Binding via custom ABI adapters. Rather than blindly rebuilding the entire execution context, the runtime tracks specific memory mutations across iterations. It categorizes each argument into one of four evaluation paths, ranging from a full descriptor allocation (on the first step) to a trivial base-pointer update or complete reuse (on steady-state steps).

By identifying that the vast majority of tensors (e.g., weights, static buffers) remain topologically identical across steps, the runtime channels nearly all arguments through zero-overhead reuse paths. Hot-loop CPU dispatch is fundamentally reduced to a handful of integer comparisons and pointer swaps, ensuring the GPU remains fully saturated.

\noindent\textbf{2. Memory-Bound Operators.}

Normalization, activation and loss kernels are limited by memory bandwidth rather than arithmetic, so the question is not which implementation is fastest in isolation but how close each comes to saturating the device. We measure LayerNorm ($16 \times 1024 \times 768$), GELU ($16 \times 1024 \times 3072$) and fused softmax cross-entropy ($16 \times 1024 \times 50304$), forward and backward, against PyTorch's Triton-backed compile mode, XLA, and Liger~\cite{liger}, a library of hand-written Triton kernels for exactly these operators. Liger provides no GELU kernel, so that comparison is three-way. Bandwidth is computed from the bytes each operator must move: one read and one write of the activation for LayerNorm and GELU, a single read of the logits for the cross-entropy forward, and a read plus a write of the vocabulary-sized gradient for its backward.

Figure~\ref{fig:membw} reports achieved bandwidth. Nova is in the fastest group on four of the six measurements, with the differences inside that group at or below measurement noise. LayerNorm forward has Nova and Liger both near 1150\,GB/s, roughly 40\% ahead of PyTorch and XLA; both exceed the card's 960\,GB/s DRAM peak because the 50\,MB activation is largely resident in its 96\,MB L2. GELU forward is a three-way tie near 800\,GB/s. GELU backward places Nova and PyTorch together at about 800\,GB/s against XLA's 374, and the cross-entropy forward places Nova, PyTorch and Liger together near 880\,GB/s, again with XLA at half.

The two gaps that are not noise are both backward passes, and both are against the hand-written kernels. LayerNorm backward reaches 730\,GB/s against Liger's 877, and the cross-entropy backward reaches 547\,GB/s against Liger's 809 and PyTorch's 792. The latter is the widest gap in our evaluation. Both implementations move the same 6.6\,GB of logical traffic, so the difference lies in how efficiently that traffic is issued rather than in how much of it there is. The two gaps do not matter equally: as Figure~\ref{fig:memlat} shows, the cross-entropy pair costs two orders of magnitude more per step than LayerNorm, so its backward is the one with end-to-end consequence.

% Fig. 11 is declared here, with the attention discussion, so it lands on
% p12 rather than crowding onto p11 with Figs. 9 and 10.
\begin{figure*}[!t]
  \centering
  {\footnotesize\swatch{novacol}~Nova\quad\swatch{ptcol}~PyTorch\quad\swatch{tritoncol}~Triton (via PyTorch)}\\[0.5em]
  \includegraphics[height=1.82in]{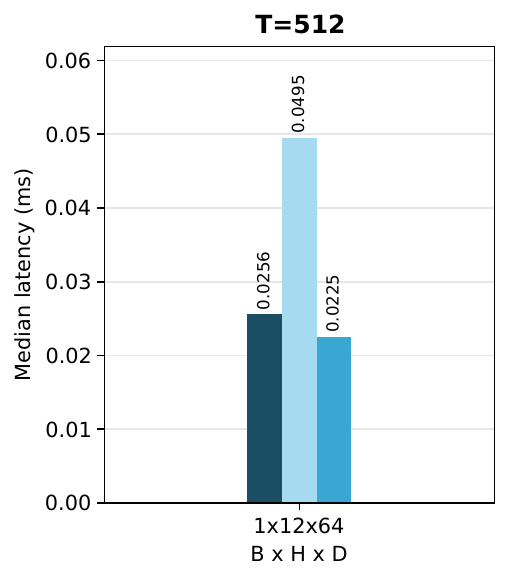}\hfill
  \includegraphics[height=1.82in]{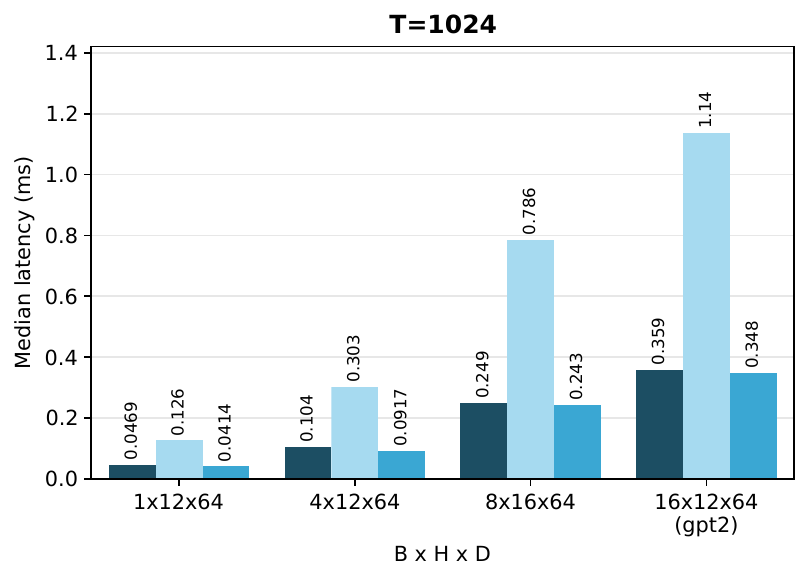}\hfill
  \includegraphics[height=1.82in]{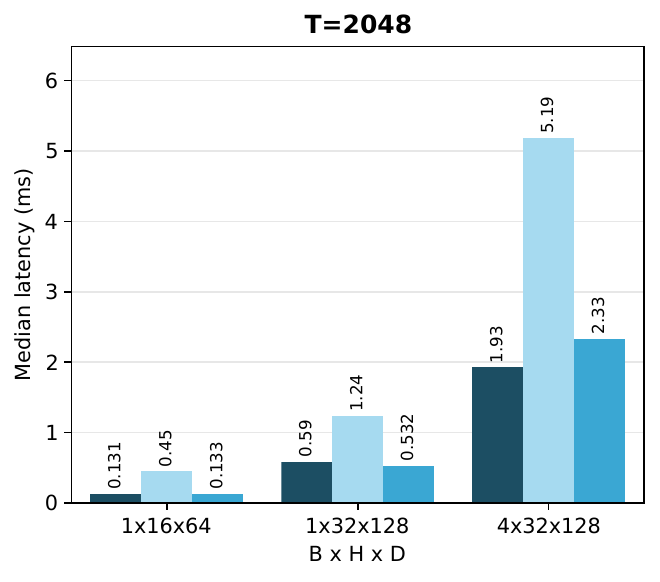}\\[0.6em]
  \includegraphics[height=1.82in]{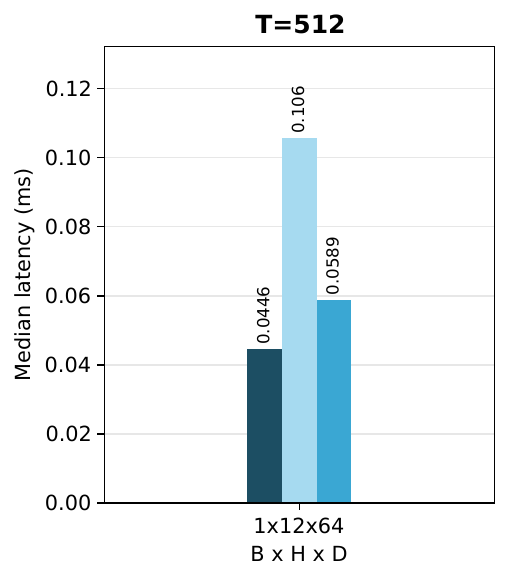}\hfill
  \includegraphics[height=1.82in]{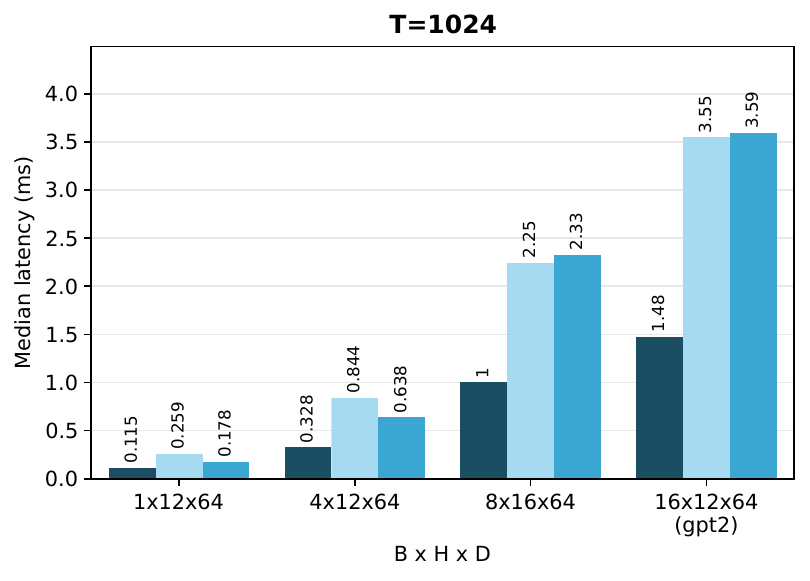}\hfill
  \includegraphics[height=1.82in]{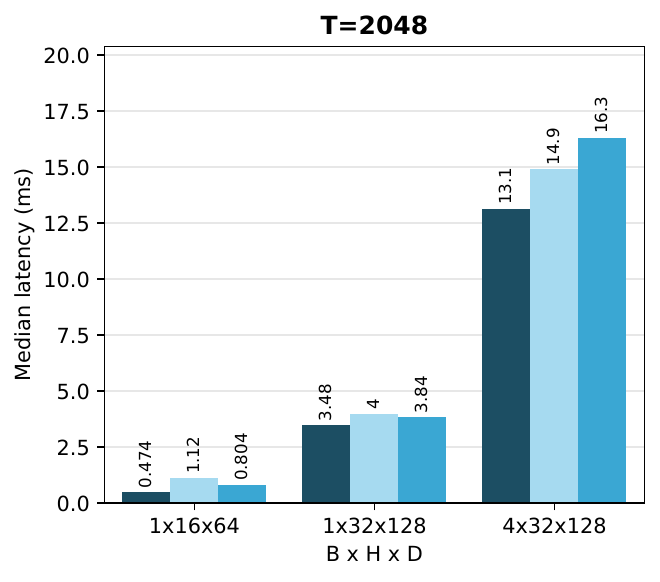}
  \caption{\textit{Causal attention latency, lower is better. Top row: forward pass. Bottom row: backward pass. Columns are sequence lengths $T = 512$, $1024$ and $2048$. The GPT-2 configuration, $16 \times 12 \times 64$, is marked in the $T = 1024$ column.}}
  \label{fig:attnbench}
\end{figure*}

\noindent\textbf{3. Attention.}

Attention is the operator most often delegated to a hand-written library kernel, which makes it the sharpest test of whether generated code can compete. We measure causal attention forward and backward at sequence lengths 512, 1024 and 2048, across head counts from 12 to 32 and head dimensions of 64 and 128, against PyTorch and against a hand-written Triton flash-attention kernel. Figure~\ref{fig:attnbench} reports the measured latencies.

On the forward pass Nova is within 14\% of the hand-written kernel on every shape and faster on two of eight, including the largest configuration measured, where it leads by 18\%. Against PyTorch it is between 1.9$\times$ and 3.4$\times$ faster throughout.

On the backward pass Nova is faster than both baselines on every shape, by 1.3$\times$ to 2.4$\times$ against Triton and by 2.2$\times$ to 2.6$\times$ against PyTorch. The advantage widens with size: at the smallest configuration it is 1.3$\times$, and at the GPT-2 configuration 2.4$\times$.

This follows from the lowering described in Section~\ref{sec:attention}. Nova compiles the backward into two fused kernels that recompute the probability matrix inside the kernel from the log-sum-exp cached by the forward pass, so the score matrix is never written to or read from memory in either direction. The hand-written kernel round-trips it.

The backward pass dominates a training step, accounting for roughly 80\% of the combined attention cost here, so this is the more consequential half. For the configuration this paper trains, 12 heads of dimension 64 at a 1024-token context, forward and backward together cost 1.84\,ms for Nova against 3.94\,ms for the Triton kernel and 4.68\,ms for PyTorch: advantages of 2.1$\times$ and 2.5$\times$. A compiler-generated attention kernel is therefore faster than a hand-written one where it matters most.

\noindent\textbf{4. Fused Loss Head.}

While the preceding measurements treat each operator in isolation, this evaluation analyzes a cross-operator fusion: the language-model head's projection combined directly with its softmax cross-entropy loss, compiled as a single kernel rather than dispatched as three. (Note that this is distinct from the cross-entropy row in the memory-bound operators above, which evaluates the loss operator purely on its own.) We report this fused execution across both hardware targets in Table~\ref{tab:losshead}. Because peak memory is dictated strictly by the problem shape rather than the underlying device, it is reported once.

Nova compiles the entire chain into a single contraction that accumulates the softmax statistics directly in its epilogue, followed by two negligible merges over the per-tile partials. In contrast, PyTorch eagerly dispatches a standard contraction, a full log-softmax pass over the materialized logits, and a negative log-likelihood reduction. As a result, Nova completes the chain 1.26$\times$ faster on the Ada 6000 and 1.55$\times$ faster on the RTX 3060.

\begin{table}[!tb]
\centering
\footnotesize
\setlength{\tabcolsep}{4pt}
\begin{tabular}{@{}llrrr@{}}
\toprule
 & \textbf{kernel} & \textbf{Ada} & \textbf{3060} & \textbf{peak} \\
 & & \textbf{(ms)} & \textbf{(ms)} & \textbf{(GB)} \\
\midrule
\multirow{3}{*}{Nova} & fused GEMM + statistics & 7.709 & 27.84 & \\
 & partial merges & 0.197 & 1.30 & \\
 & \textbf{total} & \textbf{7.91} & \textbf{29.14} & \textbf{1.63} \\
\addlinespace
\multirow{4}{*}{\makecell[l]{PyTorch\\eager}} & library GEMM & 5.830 & 25.29 & \\
 & log-softmax & 4.138 & 19.53 & \\
 & NLL reduction & 0.025 & 0.18 & \\
 & \textbf{total} & 9.99 & 45.16 & 3.07 \\
\bottomrule
\end{tabular}
\caption{\textit{Fused language-model head: projection plus softmax cross-entropy, decomposed by kernel. Peak memory depends only on the problem shape and is reported once.}}
\label{tab:losshead}
\end{table}

The execution decomposition clearly shows that this advantage does not stem from a faster contraction. On the Ada 6000, Nova's fused kernel requires 7.71\,ms, whereas the highly-tuned library kernel computes the projection alone in 5.83\,ms. Therefore, accumulating the statistics inline incurs a 32\% penalty on the contraction itself. However, this penalty buys the complete removal of the 4.14\,ms log-softmax traversal, which simply no longer exists as a standalone kernel.

Evaluating across the two hardware targets illustrates how this trade-off is bound by device architecture. Moving from the RTX 3060 to the Ada 6000, the duration of the memory traversal that Nova eliminates shrinks by a factor of 4.7$\times$ (from 19.53\,ms down to 4.14\,ms). However, the added cost of the epilogue shrinks by only a factor of 1.4$\times$ (from 2.55\,ms to 1.88\,ms). Consequently, the relative penalty on the contraction rises from 10\% to 32\%, narrowing the net speedup advantage from 1.55$\times$ to 1.26$\times$. This is consistent with the underlying argument: eliminating a memory traversal yields the highest dividends on the device with the least memory bandwidth to spare.

The impact of this fusion is equally visible in the memory footprint. Nova peaks at just 1.63\,GB compared to 3.07\,GB for standard eager execution. This disparity occurs because eager execution is forced to hold both the logits and the log-softmax output simultaneously over the identical $8192 \times 50304$ extent. Nova, conversely, holds only the logits, which the backward pass requires. The statistics themselves are negligible in size: merely tens of megabytes of per-tile partials rather than a second, full-vocabulary tensor.

\begin{table}[!b]
\centering
\footnotesize
\begin{tabular}{@{}lr@{}}
\toprule
\multicolumn{2}{@{}c@{}}{\textbf{Model \& Training}} \\
\midrule
Model size & 124M \\
$d_{\mathrm{model}}$ & 768 \\
Layers & 12 \\
Heads & 12 \\
Head dim & 64 \\
FFN hidden & 3072 \\
Non-linear activations & GELU \\
Weight tying & On \\
Context length & 1024 \\
Vocabulary & 50,257 (50,304 padded) \\
Global batch & 524,288 \\
Max learning rate & $6 \times 10^{-4}$ \\
Min learning rate & $6 \times 10^{-5}$ \\
Warmup steps & 715 \\
Training steps & 19,073 \\
Grad.\ accumulation & 4 \\
Dataset & 10B FineWeb-Edu~\cite{fineweb} \\
Parallelism & DDP, $8\times$ RTX 6000 Ada \\
\bottomrule
\end{tabular}
\caption{\textit{GPT-2 124M model and training configuration.}}
\label{tab:config}
\end{table}

% End-to-end floats are declared together, ahead of the fused-loss-head
% subsection, so the output routine releases them onto pages 13-14 beside
% their discussion in Sec. 6.2.
% They are in reading order (Figs. 12-15); do not reorder.
\begin{figure}[!b]
  \centering
  \includegraphics[width=\columnwidth]{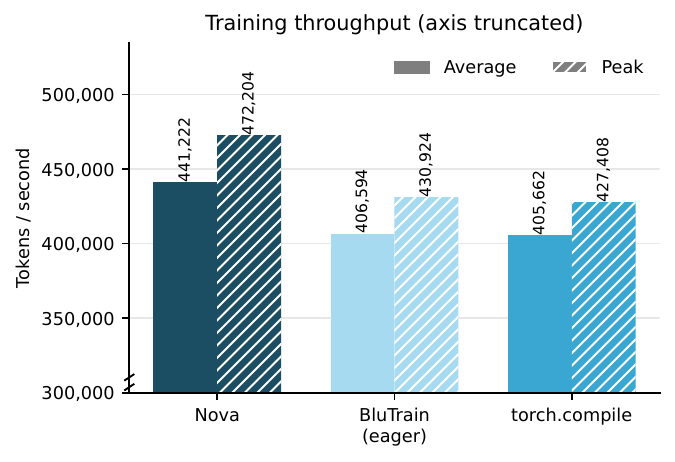}
  \caption{\textit{Sustained training throughput, aggregated across the eight ranks. Solid bars are the run average, hatched bars the peak.}}
  \label{fig:e2ethr}
\end{figure}

\begin{figure}[!b]
  \centering
  {\footnotesize\swatch{novaline}~Nova\quad\swatch{eagerline}~eager\quad\swatch{ptline}~\texttt{torch.compile}}\\[0.4em]
  \includegraphics[width=\columnwidth]{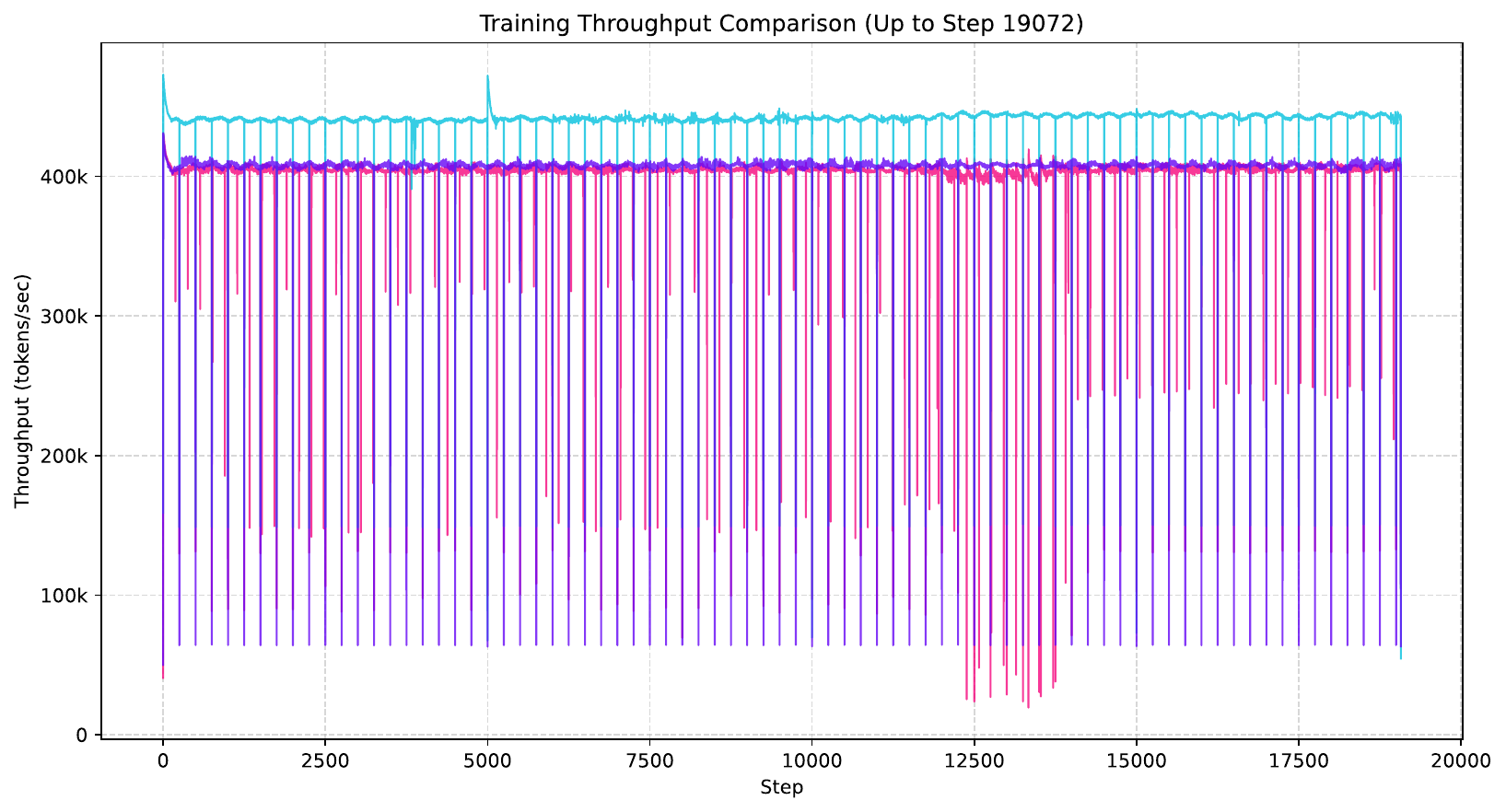}
  \caption{\textit{Sustained training throughput over the full 19,073-step run for the three execution paths. The periodic downward spikes are steps that also carry a validation or evaluation pass, whose cost is included in the step time.}}
  \label{fig:throughput}
\end{figure}

\begin{figure}[!b]
  \centering
  \includegraphics[width=0.9\columnwidth]{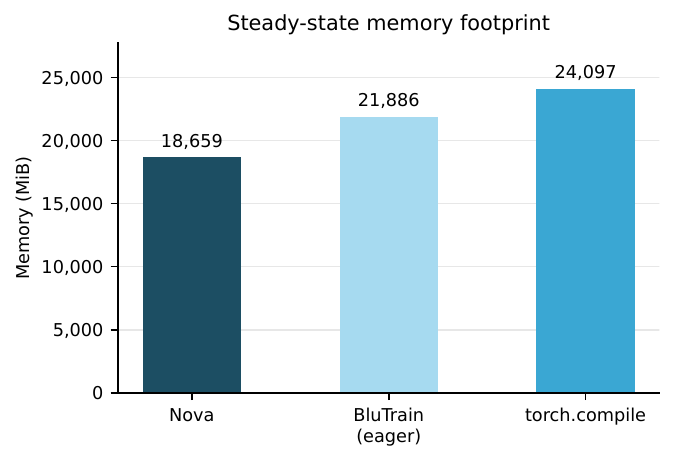}
  \caption{\textit{Steady-state device memory per rank. Nova's whole-step fusion removes the intermediates that per-operator dispatch materialises.}}
  \label{fig:e2emem}
\end{figure}

\begin{figure*}[!tb]
  \centering
  \includegraphics[width=\textwidth]{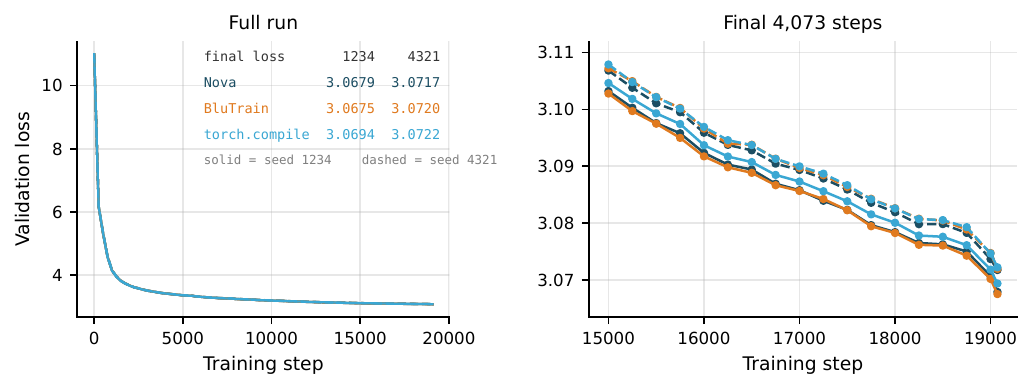}
  \caption{\textit{Validation loss for the three execution paths at two random seeds. Left: the full 19,073-step run, over which the curves are indistinguishable. Right: the final 4,073 steps on an expanded scale, where the three implementations stay within 0.002 of each other inside each seed band. Solid lines are seed 1234, dashed seed 4321; final losses are inset.}}
  \label{fig:valloss}
\end{figure*}

\subsection{End-to-End Training}\label{sec:e2e}

We train GPT-2 124M across eight RTX 6000 Ada GPUs under data-parallel training, comparing Nova against BluTrain's eager execution~\cite{blutrain} and \texttt{torch.compile}~\cite{pytorch2}. All three paths use identical model, data and batch configuration, given in Table~\ref{tab:config}. Throughput is sustained tokens per second aggregated across the eight ranks; memory is steady-state device memory per rank. Validation loss is reported at two random seeds. Figures~\ref{fig:e2ethr} and~\ref{fig:throughput} give the throughput results and the trace over the run, Figure~\ref{fig:e2emem} the memory results, and Figure~\ref{fig:valloss} the validation-loss curves.

\noindent\textbf{1. Throughput.} As Figure~\ref{fig:e2ethr} shows, Nova sustains the highest throughput of the three paths, at 441K tokens/second average and 472K peak, an advantage of 8.5\% over our own eager execution and 8.8\% over \texttt{torch.compile} in the average case, and 9.6\% and 10.5\% at peak. This figure is measured on a distributed step: because gradient synchronization overlaps computation (Section~\ref{sec:ddp}), the aggregate figure does not separate kernel improvements from communication scheduling.

\noindent\textbf{2. Memory.} Nova achieves the lowest steady-state memory footprint (Figure~\ref{fig:e2emem}), demonstrating a 14.7\% reduction compared to our eager baseline and a 22.6\% reduction versus \texttt{torch.compile}. This efficiency stems directly from our whole-step graph compilation approach. Intermediate tensors that are typically materialized by per-operator dispatch are either entirely eliminated via aggressive cross-operator fusion (Section~\ref{sec:fusion}), or avoided completely through buffer aliasing where outputs recycle the memory of dead inputs.

\noindent\textbf{3. Numerical Parity.} Nova's fusions leave convergence unchanged, as Figure~\ref{fig:valloss} shows. At seed 1234 the final validation losses are 3.0679 for Nova, 3.0675 for eager and 3.0694 for \texttt{torch.compile}; at seed 4321 they are 3.0717, 3.0720 and 3.0722. No implementation is consistently ahead: the eager baseline reaches the lowest final validation loss at one seed and Nova at the other, while \texttt{torch.compile} has the highest loss at both. Nova differs from eager by 0.0004 and 0.0002 at the two seeds and from \texttt{torch.compile} by 0.0015 and 0.0005, and the largest difference between any two implementations is 0.0019. Nova alone varies by 0.0038 between the two seeds. Differences attributable to the compiler are therefore smaller than the run-to-run variation caused by initialization. The expanded panel makes the ordering visible: the two seeds separate into distinct bands, and within each band the three implementations interleave rather than resolving into a consistent ranking.

\section{Discussions and limitations}

In this section, we analyze Nova's architectural trade-offs, evaluating the limits of its performance characteristics and open challenges for scaling.

\subsection{Empirical Optimality Gap in Analytic Scheduling}

By replacing autotuning with analytic scheduling, Nova reduces schedule selection from hours to milliseconds, but introduces a measurable optimality gap (Section~\ref{sec:microbench}). Nova trails cuBLAS by 15--21\% on the two smaller square shapes and Triton by 24\% on the batch-4 MLP projection, and achieves 68\% of Liger's bandwidth on the cross-entropy backward pass. Two structural limitations explain those gaps, and a third constrains retargeting:

\begin{itemize}
  \item \textbf{Microarchitectural non-linearities.} Static models ignore dynamic effects like L2 queuing or instruction-cache pressure, occasionally yielding suboptimal tile selections.
  \item \textbf{Unmodelled budget terms.} An incomplete cost model leads to constant fallbacks. The fused attention tile is currently pinned at $64 \times 32$ because the shared-memory budget does not yet model the driver's per-block reserve (Section~\ref{sec:attention}).
  \item \textbf{Co-design overhead.} Retargeting to new microarchitectures (e.g., Hopper TMA, AMD CDNA) requires manually extending the configurator's analytic equations.
\end{itemize}

\subsection{Graph Dynamism and JIT Latency}

Nova achieves zero execution overhead on static graphs via structural hashing. However, highly dynamic workloads (e.g., autoregressive generation) cause constant hash misses. This triggers JIT recompilation on the CPU hot path, introducing severe latency stalls. Supporting dynamic graphs requires future integration of symbolic shape representations and shape-generalized execution templates.

\subsection{Ecosystem Ingestion and Portability Boundaries}

Nova currently operates as a JIT backend exclusively for BluTrain. Integrating with broader frameworks like PyTorch or JAX requires robust dialect converters (e.g., via TorchDynamo) and intercepting framework-specific memory invariants without overhead. Additionally, Nova's code generation is tightly coupled to NVIDIA PTX; supporting AMD or Intel GPUs demands rewriting low-level vector lowering and swizzling passes.

\subsection{Static Communication Scheduling}

Nova achieves network-compute overlap by statically injecting all-reduce callbacks (\texttt{nova\_ddp\_bucket\_ready}) into the JIT-compiled IR. However, in large-scale deployments, network latency is highly stochastic. Because synchronization boundaries are hardcoded, thread blocks stall entirely during transient network lags, precluding the out-of-order execution enabled by dynamic runtime schedulers.

\section{Related Work}

Nova builds upon a rich lineage of deep learning compilers but fundamentally diverges in its approach to end-to-end training. While compilers like XLA~\cite{xla} excel at aggressive whole-graph fusion, they operate as shape-specializing Just-in-Time (JIT) compilers that trigger costly recompilations whenever they encounter new dynamic tensor shapes. Nova mitigates this by pairing JIT compilation with a structural hashing runtime, and natively weaves data-parallel synchronization (\texttt{nova\_ddp\_bucket\_ready}) directly into the intermediate representation~\cite{pytorchddp}.

While Nova shares structural lineage with IREE~\cite{iree}---leveraging similar MLIR abstractions for loop tiling and vectorization---IREE's production focus is inference and deployment. Nova instead centers on dynamic training, introducing native graph tracing and autograd capabilities~\cite{jax} that completely fuse the forward and backward passes.

Furthermore, Nova completely automates hardware mapping without relying on intermediate developer-facing DSLs (like Triton~\cite{triton}) or expensive search-based autotuning (like TVM and Ansor~\cite{tvm,ansor}). While developer-facing DSLs require users to manually hand-tune block sizes in Python---and while modern framework compilers like PyTorch 2.0~\cite{pytorch2} attempt to hide this complexity by auto-generating Triton code---they ultimately still rely on heuristic autotuning to discover optimal schedules. Discovering these schedules via evolutionary search can take hours and must be re-run whenever shapes change. Instead, Nova utilizes an Analytic Configurator that deterministically derives a near-optimal schedule from Arithmetic Intensity~\cite{roofline} in milliseconds, demonstrating that exhaustive search is unnecessary when hardware limits are explicitly modeled.

To ensure these analytic schedules approach physical hardware limits during training, Nova relies on an entirely custom hardware backend. While it shares high-level vectorization abstractions with IREE, Nova's low-level code generation---including elementwise fusions, software-pipelined asynchronous memory copies, XOR shared-memory swizzling, precision-aware \texttt{ldmatrix} routing~\cite{cudaguide}, and warp-level butterfly shuffles---is implemented from scratch within its own backend, drawing on techniques popularized by high-performance GEMM libraries such as CUTLASS~\cite{cutlass}. This purpose-built backend targets near-peak tensor-core utilization for dynamic training workloads without relying on closed-source libraries such as cuBLAS/cuDNN~\cite{cudnn,cublas} or search-based autotuning.

Finally, as transformer architectures have come to dominate deep learning, the ecosystem has increasingly relied on hand-written, highly specialized kernels---such as FlashAttention~\cite{flashattention} for causal masking and Liger Kernel~\cite{liger} for memory-bound LLM operators---to circumvent framework limitations. While these libraries provide peak performance, they create opaque boundaries that prevent cross-operator fusions in the broader computation graph. Nova demonstrates that by elevating operations like \texttt{online\_attention} to first-class compiler primitives (Section~\ref{sec:attention}) and managing memory hierarchies analytically, a single unified compilation pipeline can achieve parity with these bespoke kernels while retaining full end-to-end graph optimization.

\section{Conclusion and Future Work}

In this paper we presented Nova, an end-to-end compiler that turns a complete training step, forward and backward alike, into fused machine code, deriving how to run it from the structure of the computation rather than from a library of pre-written kernels. Because it works from structure, Nova compiles familiar and novel operations through the same pipeline, with no hand-tuned kernel written for either and no retuning when the target hardware changes.

This iteration extends that pipeline to full transformer training. Attention is compiled rather than called: the score matrix is never materialised in either direction, causality is exploited as loop structure rather than as data, and the resulting kernel is faster than a hand-written flash-attention kernel on the backward pass that dominates training. Across matrix multiplication, normalization, activation and loss kernels, generated code reaches or exceeds vendor libraries and hand-written Triton on the shapes that dominate a language-model step, at comparable numerical accuracy, and a compiled step holds a 15 to 25\% lower memory footprint than either at equal configuration. The same step extends without change to data-parallel training across eight GPUs.

\subsection*{Future Work}

Nova compiles a complete transformer training step for NVIDIA hardware. Because the schedule is computed entirely from device parameters read at compile time, and hardware specialization is confined to the final code-generation stage, the directions below broaden coverage while leaving the analytic core intact.

\noindent\textbf{Dynamic shapes.} Structural hashing gives zero steady-state overhead on a static graph, but recompiles whenever a shape changes, which rules out autoregressive generation. Symbolic shape representations and shape-generalized execution templates would remove that limit, and a persistent artifact cache would remove the per-process compilation cost.

\noindent\textbf{Wider architecture support.} We plan to retarget Nova to newer NVIDIA generations such as Hopper and Blackwell, and to non-NVIDIA hardware including AMD GPUs and multicore CPUs. Extending coverage requires a new hardware description and new low-level vector lowering and swizzling passes, not a new scheduling strategy.

% Wired bibliography: numbers auto-assigned in listing order (1--23 today).

\end{document}